\documentclass{article}

\usepackage[preprint]{neurips_2023}

\usepackage{hyperref}

\usepackage[utf8]{inputenc} 
\usepackage[T1]{fontenc}    
\usepackage{hyperref}       
\usepackage{url}            
\usepackage{booktabs}       
\usepackage{amsfonts}       
\usepackage{nicefrac}       
\usepackage{microtype}      
\usepackage{xcolor}         

\usepackage{amsmath,amsfonts,bm}

\def\eqref#1{equation~\ref{#1}}

\def\1{\bm{1}}

\def\rmG{{\mathbf{G}}}
\def\rmH{{\mathbf{H}}}

\def\rmL{{\mathbf{L}}}
\def\rmM{{\mathbf{M}}}

\def\rmR{{\mathbf{R}}}
\def\rmS{{\mathbf{S}}}

\def\rmW{{\mathbf{W}}}
\def\rmX{{\mathbf{X}}}

\def\rmZ{{\mathbf{Z}}}

\def\ermG{{\textnormal{G}}}

\def\ermZ{{\textnormal{Z}}}

\def\vzero{{\bm{0}}}

\def\vmu{{\bm{\mu}}}

\def\ve{{\bm{e}}}

\def\vh{{\bm{h}}}

\def\vu{{\bm{u}}}
\def\vv{{\bm{v}}}
\def\vw{{\bm{w}}}
\def\vx{{\bm{x}}}
\def\vy{{\bm{y}}}
\def\vz{{\bm{z}}}

\def\evu{{u}}
\def\evv{{v}}
\def\evw{{w}}
\def\evx{{x}}
\def\evy{{y}}
\def\evz{{z}}

\def\mH{{\bm{H}}}
\def\mI{{\bm{I}}}

\def\mZ{{\bm{Z}}}

\def\mPhi{{\bm{\Phi}}}

\def\mPhi{{\bm{\Phi}}}

\DeclareMathAlphabet{\mathsfit}{\encodingdefault}{\sfdefault}{m}{sl}
\SetMathAlphabet{\mathsfit}{bold}{\encodingdefault}{\sfdefault}{bx}{n}
\newcommand{\tens}[1]{\bm{\mathsfit{#1}}}

\def\tT{{\tens{T}}}

\def\tW{{\tens{W}}}
\def\tX{{\tens{X}}}

\def\gL{{\mathcal{L}}}

\def\gN{{\mathcal{N}}}

\def\sC{{\mathbb{C}}}

\def\sN{{\mathbb{N}}}

\def\sR{{\mathbb{R}}}
\def\sS{{\mathbb{S}}}

\newcommand{\etens}[1]{\mathsfit{#1}}

\def\etT{{\etens{T}}}

\def\etW{{\etens{W}}}

\def\etZ{{\etens{Z}}}

\newcommand{\R}{\mathbb{R}}

\newcommand{\supp}{\mathrm{Supp}}

\DeclareMathOperator*{\argmax}{arg\,max}
\DeclareMathOperator*{\argmin}{arg\,min}

\DeclareMathOperator{\sign}{sign}
\DeclareMathOperator{\Tr}{Tr}

\DeclareMathOperator{\asto}{\xrightarrow{\text{a.s.}}}
\DeclareMathOperator{\toind}{\xrightarrow{\mathcal{D}}}
\newcommand{\spec}{\mathrm{Sp}}
\newtheorem{theorem}{Theorem}
\newtheorem{proposition}{Proposition}
\newtheorem{lemma}{Lemma}
\newtheorem{assumption}{Assumption}
\newtheorem{remark}{Remark}
\usepackage{algorithm}
\usepackage{algpseudocode}
\usepackage{wrapfig}

\usepackage{cancel}
\usepackage{pgfplots}
\pgfplotsset{compat = newest}
\usepackage{tikz}
\usepackage{pgfplots}
\usetikzlibrary{matrix}
\usepgfplotslibrary{groupplots}
\pgfplotsset{compat=newest}
\pgfplotsset{width=7.5cm,compat=1.12}
\usepgfplotslibrary{fillbetween}
\usepackage{varioref}

\usepackage{cleveref}
\usepackage{comment}

\definecolor{islamicgreen}{RGB}{0, 144, 81}
\definecolor{myblue}{RGB}{0, 144, 240}

\hypersetup{colorlinks,linkcolor={purple},citecolor={purple},urlcolor={purple}} 

\title{A Nested Matrix-Tensor Model \\ for Noisy Multi-view Clustering}

\author{%
  Mohamed El Amine Seddik\And Mastane Achab\And Henrique Goulart\And Merouane Debbah
}

\begin{document}

\maketitle

\begin{abstract}
In this paper, we propose a nested matrix-tensor model which extends the spiked rank-one tensor model of order three.
This model is particularly motivated by a multi-view clustering problem in which multiple noisy observations of each data point are acquired, with potentially non-uniform variances along the views. In this case, data can be naturally represented by an order-three tensor where the views are stacked. 
Given such a tensor, we consider the estimation of the hidden clusters via performing a best rank-one tensor approximation.
In order to study the theoretical performance of this approach, we characterize the behavior of this best rank-one approximation in terms of the alignments of the obtained component vectors with the hidden model parameter vectors, in the large-dimensional regime.
In particular, we show that our theoretical results allow us to anticipate the exact accuracy of the proposed clustering approach. Furthermore, numerical experiments indicate that leveraging our tensor-based approach yields better accuracy compared to a naive unfolding-based algorithm which ignores the underlying low-rank tensor structure.
Our analysis unveils unexpected and non-trivial phase transition phenomena depending on the model parameters, ``interpolating'' between the typical behavior observed for the spiked matrix and tensor models.
\end{abstract}

\section{Introduction}

Tensor methods have received growing attention
in recent years, especially from a statistical perspective, following the introduction of a statistical model for tensor PCA by \cite{montanari2014statistical}. 
In machine learning, these methods are particularly attractive for addressing several unsupervised learning tasks which can be formulated as the extraction of some \emph{low-rank structure} from a (potentially high-dimensional) tensor containing observations or functions thereof (such as high-order moments).
Among the many existing examples, we can mention learning latent variable models such as Dirichlet allocation, topic models, multi-view models and Gaussian mixtures \citep{anandkumar2014tensor, AnanGJ-15-PMLR, GeHK-15-STC, HsuKZ-12-JCSC, HsuK-13-CITCS, JanzGKA-19-FTML, KhouMM-22-JSC, BaksDJK-22-STC, RahmNFT-20-PRL}; learning probability densities and non-Gaussian mixtures \citep{KargS-19-PMLR, SingMRG-23-arxiv, OselK-21-CMMP}; detecting communities from interaction data of (possibly multi-view or time-evolving) networks \citep{AnanGHK-13-PMLR, HuanNH-15-JMLR, GujrPP-20-TWC, FernFG-21-AIR}; and high-order co-clustering \citep{PapaSB-12-TSP}.


Despite its simplicity, the statistical model of \cite{montanari2014statistical}, sometimes called a \textit{rank-one spiked tensor model}, has raised many theoretical challenges. A significant amount of work has been done to understand the fundamental questions related to this model \citep{perry2016statistical, jagannath2020statistical,  goulart2021random, auddy2022estimating, arous2021long, seddik2021random}, in particular involving statistical thresholds and the asymptotic performance of estimators in the large-dimensional limit.
However, the findings of these works have a somewhat limited practical impact due to the rank-one nature of that model, motivating the development and study of more sophisticated statistical models for the analysis of tensor methods. 
In particular, phase transitions associated with multi-spiked tensor models of rank $r > 1$ have been considered by \cite{chen2021phase, lesieur2017statistical}.

In this work, we take another path towards bridging the gap between theory and practical applications, by proposing a statistical \textit{nested matrix-tensor model} that generalizes the (third-order) rank-one spiked tensor model and is motivated by a problem that we call \emph{noisy multi-view clustering}, which can be formulated as follows.
Let $\rmM = \vmu \vy^\top + \rmZ \in \R^{p \times n}$ be a data matrix containing $n$ observations of $p$-dimensional vectors centered around $\pm \vmu$ (i.e., data are made of two classes), with $\vy \in \{ -1, 1\}^n$ holding their corresponding labels and $\rmZ$ a Gaussian matrix modeling data dispersion.
Now, suppose that we are given $m$ different noisy observations of $\rmM$ with potentially different signal-to-noise ratios (SNR), denoted by:
\[
  \tilde{\rmX}_k = \vmu \vy^\top + \rmZ + \tilde{\rmW}_k, \quad k=1,\ldots,m,
\]
where $\tilde\rmW_k$ is a $p \times n$ matrix comprising independent Gaussian entries drawn from $\mathcal{N}(0,\sigma_k^2)$.
Assuming that the variances $\sigma_k^2$ are known (or can be accurately estimated), one can build a tensor $\tX \in \R^{p \times n \times m}$ containing normalized slices $\rmX_k = h_k \tilde{\rmX}_k$, with $h_k := 1/\sigma_k$, so that: 
\begin{align}\tag{Nested Matrix-Tensor Model}\label{eq:intro_nested}
    \tX = \left( \vmu \vy^\top + \rmZ \right) \otimes \vh + \tW \, \in \sR^{p\times n \times m},
\end{align}
where the tensor $\tW$ has independent standard Gaussian entries and $\vh=(h_1,\ldots,h_m)^\top \in \sR^m$.


The above model can be seen as a more general version of the rank-one spiked model that incorporates a nested structure allowing for more flexible modeling (Specifically, when the variances of the elements in $\rmZ$ tend to zero, one recovers the rank-one spiked model).
The common low-rank structure in the slices $\rmX_k$, which can be interpreted as different views of the data, encodes the latent clustering structure that can then be retrieved by using tensor methods applied on $\tX$.

In particular, our results precisely quantify the asymptotic performance of a simple estimator of the vectors $\vmu, \vy$, and $\vh$ based on rank-one approximation of $\tX$, in the large-dimensional limit where $p,n,m \to \infty$ at the same rate.
This is achieved by resorting to the recently developed approach of \cite{goulart2021random} and \cite{seddik2021random}, which allows one to use tools from random matrix theory by inspecting \emph{contractions} of the random tensor model in question.
Numerical results are given to illustrate the usefulness of such predictions even for moderately large values of $p$ and $n$, and also to show the superiority of such a tensor-based approach in comparison with a naive spectral method that does not take the tensor structure of the model into account.
Quite interestingly, our results show that the performance of such a rank-one spectral estimator exhibits different phase transition behaviors depending on two parameters governing the SNR and the data dispersion, effectively ``interpolating'' between phase transition curves that are characteristic of matrix and tensor models.


\textbf{Key contributions:} Our main contributions can be summarized as follows:
\begin{enumerate}
    \item We introduce a nested matrix-tensor model that generalizes the (third-order) spiked tensor model, and we provide a random matrix analysis of its best rank-one tensor approximation in the high-dimensional regime.
    \item We provide an application of this model to the problem of clustering multi-view data and show that the developed theory allows the exact characterization of the asymptotic performance of a multi-view clustering approach. Further simulations suggest the superiority of the tensor-based clustering approach compared to a naive unfolding method that ignores the hidden rank-one structure.
\end{enumerate}



\textbf{Related work on tensor multi-view methods:}
In multi-view machine learning \citep{xu2013survey, zhao2017multi, sun2013survey}, one has to deal with data
coming from different sources or exhibiting various statistical or physical natures (e.g. documents composed of both text and images).
The main challenge consists in jointly leveraging both the agreement and the complementarity of the different views \citep{blum1998combining,dasgupta2001pac, nigam2000analyzing}, e.g. via learning a shared latent subspace \citep{white2012convex} for diverse tasks such as regression \citep{kakade2007multi} or clustering \citep{chaudhuri2009multi,gao2015multi,cao2015diversity}.
In this context, multi-view clustering algorithms using a low-rank tensor representation of the multi-view data have already been proposed: among others, \cite{xie2018unifying,wu2020unified} relied on tensor-SVD \citep{kilmer2013third} while \cite{6193101} favored a Tucker-type tensor decomposition.

However, the usual sense employed for the term ``multi-view clustering'' is not exactly the same that we adopt here, since in our problem all views essentially hold noisy measurements of the same quantities.
Hence, our work is perhaps closer in spirit to certain tensor-based clustering models comprising an additional diversity (e.g., temporal), such as those of \cite{PapaSB-12-TSP} or those reviewed in \cite{FernFG-21-AIR}.
Yet, it differs from this literature in that our additional diversity is quite specific (namely, it comes from the availability of multiple measurements for each individual in the sample) and, furthermore, we derive the exact asymptotic performance of our proposed tensor-based method in the large-dimensional limit.



\section{Notation and Background}
The set $\{1, \ldots, n\}$ is denoted by $[n]$.
The unit sphere in $\sR^p$ is denoted by $\sS^{p-1}$.
The Dirac measure at some real value $x$ is denoted by $\delta_x$. The support of a measure $\nu$ is denoted by $\supp(\nu)$. The inner product between two vectors $\vu,\vv$ is denoted by $\langle \vu, \vv \rangle = \sum_i \evu_i \evv_i$. The imaginary part of a complex number $z$ is denoted by $\Im[z]$. The set of eigenvalues of a matrix $\rmM$ is denoted by $\spec(\rmM)$. Almost sure convergence of a sequence of random variables is denoted by $\asto$. The arrow $\toind$ denotes the convergence in distribution.

\subsection{Tensor Notations and Contractions}\label{sec_tensor_notations}
In this section, we introduce the main tensor notations and definitions used throughout the paper, which we recommend following carefully for a clear understanding of its technical contents.

\textbf{Three-order tensors:} The set of third-order tensors of size $n_1\times n_2\times n_3$ is denoted $\sR^{n_1\times n_2 \times n_3}$. The scalar $\etT_{ijk}$ or $[\tT]_{ijk}$ denotes the $(i,j,k)$ entry of a tensor $\tT\in \sR^{n_1\times n_2 \times n_3}$.

\textbf{Rank-one tensors:} A tensor $\tT$ is said to be of rank-one if it can be represented as the outer product of three real-valued vectors $(\vx,\vy,\vz)\in \sR^{n_1} \times \sR^{n_2} \times \sR^{n_3}$. In this case, we write $\tT = \vx\otimes \vy \otimes \vz$, where the outer product is defined such that $[\vx\otimes \vy \otimes \vz]_{ijk} = \evx_i \evy_j \evz_k$.

\textbf{Tensor contractions:} The first mode contraction of a tensor $\tT$ with a vector $\vx$ yields a matrix denoted $\tT(\vx,\cdot, \cdot)$ with entries $[\tT(\vx,\cdot, \cdot)]_{jk} = \sum_{i=1}^{n_1} \evx_i \etT_{ijk}$. Similarly, $\tT(\cdot, \vy, \cdot)$ and $\tT(\cdot, \cdot, \vz)$ denote the second and third mode contractions of $\tT$ with vectors $\vy$ and $\vz$ respectively. We will sometimes denote these contractions by $\tT(\vx)$, $\tT(\vy)$, and $\tT(\vz)$ if there is no ambiguity. 
The contraction of $\tT$ with two vectors $\vx,\vy$ is a vector denoted $\tT(\vx,\vy,\cdot)$ with entries $[\tT(\vx,\vy,\cdot)]_k = \sum_{ij} \evx_i \evy_j \etT_{ijk}$. Similarly, the contraction of $\tT$ with three vectors $\vx,\vy,\vz$ is a scalar denoted $\tT(\vx,\vy,\vz) = \sum_{ijk} \evx_i \evy_j \evz_k \etT_{ijk}$. 

\textbf{Tensor norms:} The Frobenius norm of a tensor $\tT$ is denoted $\Vert \tT\Vert_F$ with $\Vert \tT\Vert_F^2 = \sum_{ijk} \etT_{ijk}^2$. The spectral norm of $\tT$ is $\Vert \tT \Vert = \sup_{\Vert \vu \Vert = \Vert \vv\Vert = \Vert \vw \Vert = 1} \vert \tT(\vu,\vv,\vw) \vert $.

\textbf{Best rank-one approximation:} A best rank-one approximation of $\tT$ corresponds to a rank-one tensor of the form $\lambda \vu\otimes \vv \otimes \vw$, where $\lambda>0$ and $\vu,\vv,\vw$ are unitary vectors, that minimizes the square loss $\Vert \tT - \lambda \vu\otimes \vv \otimes \vw\Vert_F^2$. The latter generalizes to tensors the concept of singular value and vectors \citep{lim2005singular} and the scalar $\lambda$ coincides with the spectral norm of $\tT$.
Such a best rank-one approximation can be computed via \textit{tensor power iteration} which consists of iterating:
\begin{align*}
    \vu \leftarrow \tT(\cdot, \vv, \vw) / \Vert \tT(\cdot, \vv, \vw)\Vert, \quad
    \vv \leftarrow \tT(\vu, \cdot, \vw) / \Vert \tT(\vu, \cdot, \vw)\Vert, \quad
    \vw \leftarrow \tT(\vu, \vv, \cdot) / \Vert \tT(\vu, \vv, \cdot)\Vert, 
\end{align*}
starting from some appropriate initialization \citep{KofiR-02-SIMAX, anandkumar2014tensor}.

\subsection{Random Matrix Theory}
In this section, we recall some necessary tools from random matrix theory (RMT) which are at the core of our main results. Specifically, we will consider the \textit{resolvent} formalism of \cite{hachem2007deterministic} which allows one to characterize the spectral behavior of large symmetric random matrices and the estimation of low-dimensional functionals of such matrices. Given a symmetric matrix $\rmS\in \sR^{n\times n}$, the resolvent of $\rmS$ is defined as $\rmR(\xi) = \left( \rmS - \xi \mI_n \right)^{-1}$ for some $\xi\in \sC\setminus \spec(\rmS)$.

In essence, RMT focuses on describing the distribution of eigenvalues of large random matrices. Typically, under certain technical assumptions on some random matrix $\rmS\in \sR^{n\times n}$ with eigenvalues $\lambda_1, \ldots, \lambda_n$, the \textit{empirical spectral measure} of $\rmS$, defined as $\hat \nu_n = \frac1n \sum_{i=1}^n \delta_{\lambda_i}$, converges in the weak sense \citep{van1996weak} to some deterministic probability measure $\nu$ as $n\to \infty$ and RMT aims at describing such a $\nu$. To this end, one widely considered (so-called analytical) approach relies on the \textit{Stieltjes transform} \citep{widder1938stieltjes}. Given a probability measure $\nu$, the Stieltjes transform of $\nu$ is defined as $g_\nu(\xi) = \int \frac{d\nu(\lambda)}{\lambda - \xi}$ with $\xi\in \sC\setminus \supp(\nu)$, and the inverse formula allows one to describe the density of $\nu$ as $\nu(dx) = \frac{1}{\pi} \lim_{\varepsilon\to 0} \Im[g_\nu(x+i\varepsilon)]$ (assuming it admits one). 

The Stieltjes transform of the empirical spectral measure, $\hat\nu_n$, is closely related to the resolvent of $\rmS$ through the normalized trace operator. In fact, $g_{\hat\nu_n}(\xi) = \frac1n \Tr \rmR(\xi)$ and the point-wise \textit{almost sure} convergence of $g_{\hat\nu_n}(\xi)$ to some deterministic Stieltjes transform $g_\nu(\xi)$ (where $\nu$ is defined on $\R$) on the upper-half complex plane is equivalent to the weak convergence of $\hat\nu_n$ to $\nu$ \citep{tao2012topics}. Our analysis relies on estimating quantities involving $\frac1n \Tr \rmR(\xi)$, making the use of the resolvent approach a natural choice (see Appendix \ref{appendix:proofs} for the derivation of our results).

\section{Main Results}

\subsection{The Nested Matrix-Tensor Model}
We start by defining our considered nested matrix-tensor model in a general form since it might have applications beyond the multi-view data model in Eq. (\ref{eq:intro_nested}). Let $n_1, n_2, n_3 \in \sN_+$ and further denote $n_M=n_1+n_2$ and $n_T=n_1+n_2+n_3$.
We consider the following statistical model:
\begin{equation}
\label{eq:nested_model}
    \tT = \beta_T \rmM \otimes \vz + \frac{1}{\sqrt{n_T}} \tW \, \in \mathbb{R}^{n_1 \times n_2 \times n_3}, \quad
    \rmM = \beta_M \vx \otimes \vy + \frac{1}{\sqrt{n_M}} \rmZ \, \in \mathbb{R}^{n_1 \times n_2 },
\end{equation}
where we assume that $\|\vx\|=\|\vy\|=\|\vz\|=1$ and that the entries of $\tW$ and $\rmZ$ are independent Gaussian random variables, with $\etW_{ijk} \sim \gN(0, \sigma_T^2)$ and $\etZ_{ij} \sim \gN(0, \sigma_M^2)$. For the sake of simplicity, we consider the unit variance case $\sigma_T = \sigma_M = 1$ in the remainder of the paper while we defer the general variance case to Appendix \ref{appendix:proofs}.

\begin{remark}[Spectral normalization] Note that the normalization of $\tW$ by $\sqrt{n_T}$ (resp. $\mZ$ by $\sqrt{n_M}$) in Eq.~(\ref{eq:nested_model}) ensures that the spectral norm of $\tT$ is of order $O(1)$ when the dimensions $n_i$ grow to infinity. This follows from a standard concentration result \citep[Lemma 4]{seddik2021random}.
\end{remark}

\textbf{Best rank-one tensor estimator:} We consider the analysis of the best rank-one approximation of $\tT$ which corresponds to the following problem \citep{lim2005singular}:
\begin{equation}
\label{eq:deflation}
    \argmin_{\lambda>0,\, \|\vu\| \|\vv\| \|\vw\|=1} \| \tT - \lambda \vu \otimes \vv \otimes \vw \|_{\text{F}}^2 \quad
    \Leftrightarrow \quad \argmax_{\|\vu\| \|\vv\| \|\vw\|=1} \tT(\vu, \vv, \vw) \ .
\end{equation}
In particular, the solution for the scalar $\lambda$ in the left-hand problem coincides with the spectral norm of $\tT$, i.e., $\lambda = \|\tT\|$.
Given a critical point $(\lambda, \vu, \vv, \vw)$ of that problem,
it holds that \citep{lim2005singular}:
\begin{equation}
\label{eq:eigen}
    \tT(\cdot, \vv, \vw) = \lambda \vu, \quad
    \tT(\vu, \cdot, \vw) = \lambda \vv, \quad
    \tT(\vu, \vv, \cdot) = \lambda \vw, \quad \lambda = \tT(\vu, \vv, \vw).
\end{equation}

In essence, for sufficiently large $\beta_M$ and $\beta_T$, the triplet $(\vu, \vv, \vw)$ will start to align with the signal components $(\vx, \vy, \vz)$ and our main goal is to quantify these alignments (i.e., the inner products $\langle \vu, \vx \rangle$, $\langle \vv, \vy \rangle$ and $\langle \vw, \vz \rangle$) in the large dimensional regime when $n_i\to \infty$. To this end, we need a typical set of assumptions that we formulate as follows (see \citep{goulart2021random, seddik2021random} for similar assumptions in the case of spiked random tensors).
\begin{assumption}
\label{ass:convergence}
    There exists a sequence of critical points $(\lambda, \vu, \vv, \vw)$ satisfying Eq. (\ref{eq:eigen}) such that,
    when $n_i \to \infty$ with $\frac{n_1}{n_T} \to c_1 > 0, \frac{n_2}{n_T} \to c_2 > 0, \frac{n_3}{n_T} \to c_3 > 0$, we have the following:
    \begin{equation*}
    \lambda \asto \bar{\lambda} , \quad
    \vert\langle \vu, \vx \rangle\vert \asto \alpha_1 , \quad
    \vert\langle \vv, \vy \rangle\vert \asto \alpha_2 , \quad
    \vert\langle \vw, \vz \rangle\vert \asto \alpha_3.
    \end{equation*}
\end{assumption}
In the remainder of the paper, we refer to the quantities $(\lambda,\langle \vu, \vx \rangle,\langle \vv, \vy \rangle,\langle \vw, \vz \rangle)$ as \textit{summary statistics} as per the formalism introduced by \cite{ben2022high} since the asymptotic limits of these scalar quantities fully describe the asymptotic behavior of the considered best rank-one tensor estimator applied to $\tT$.

\begin{remark}[On Assumption \ref{ass:convergence}]
The almost sure convergence of the summary statistics has been demonstrated in \citep{jagannath2020statistical} in the case of the spiked tensor model. We believe similar arguments can be extended to our proposed nested matrix-tensor model to validate Assumption \ref{ass:convergence}.
\end{remark}

\subsection{Associated Random Matrix}
As discussed in the previous section, our primary goal is to compute the asymptotic summary statistics $(\bar \lambda, \alpha_1, \alpha_2, \alpha_3)$ in terms of the model's parameters, namely, the signal-to-noise ratios $(\beta_M, \beta_T)$ and the dimension ratios $(c_1, c_2, c_3)$.
To this end, we follow the approach developed by \cite{seddik2021random}, who studied the \textit{asymmetric} spiked tensor model, and where it has been shown that the estimation of $(\bar \lambda, \alpha_1, \alpha_2, \alpha_3)$ boils down to the analysis of the \textit{block-wise contraction random matrix} $\mPhi$ in Eq.(\ref{eq:def_phi}), which can be done by deploying tools from random matrix theory.
\begin{wrapfigure}{r}{0.4\textwidth}
  \begin{small}
    \begin{align}\label{eq:def_phi}
    \mPhi = \begin{bmatrix}
        \vzero_{n_1\times n_1} & \tT(\vw) & \tT(\vv) \\
        \tT(\vw)^\top & \vzero_{n_2\times n_2} & \tT(\vu) \\
        \tT(\vv)^\top & \tT(\vu)^\top & \vzero_{n_3\times n_3}
    \end{bmatrix}
\end{align}
\end{small}
\end{wrapfigure}

Given the model in Eq. (\ref{eq:nested_model}), it can be easily noticed that $\mPhi$ decomposes as a sum of two matrices $\rmH + \rmL$ where $\rmL$ is a low-rank matrix related to the signal part in the nested matrix-tensor model (the expression of $\rmL$ is provided in Eq. (\ref{eq:expression_of_L}) in Appendix \ref{appendix:proofs}), and $\rmH$ corresponds to the noise part of the model, being given by:
\begin{small}
    \begin{equation}
    \rmH = \begin{bmatrix}
    \vzero_{n_1\times n_1} & \frac{ \langle \vw, \vz \rangle \beta_T }{ \sqrt{n_M} } \rmZ + \frac{1}{\sqrt{n_T}} \tW(\vw) & \frac{1}{\sqrt{n_T}} \tW(\vv) \\
    \frac{ \langle \vw, \vz \rangle \beta_T }{ \sqrt{n_M} } \rmZ^\top + \frac{1}{\sqrt{n_T}} \tW(\vw)^\top & \vzero_{n_2\times n_2} & \frac{1}{\sqrt{n_T}} \tW(\vu) \\
    \frac{1}{\sqrt{n_T}} \tW(\vv)^\top & \frac{1}{\sqrt{n_T}} \tW(\vu)^\top & \vzero_{n_3\times n_3}
    \end{bmatrix} .
\end{equation}
\end{small}

\begin{remark}[On the spectrum of $\mPhi$]\label{remark:spikes}
    In terms of spectrum, we will see subsequently that the matrices $\mPhi$ and $\rmH$ share the same ``bulk'' of eigenvalues while the spectrum of $\mPhi$ exhibits two isolated eigenvalues at positions $2\lambda$ and $-\lambda$ if $\beta_M, \beta_T$ are large enough. In fact, one can quickly check that, given the identities in Eq. (\ref{eq:eigen}), the scalars $2\lambda$ and $-\lambda$ are eigenvalues of $\mPhi$ with respective multiplicities $1$ and $2$, and respective eigenvectors $(\vu^\top, \vv^\top, \vw^\top)^\top$ for the eigenvalue $2\lambda$
    and $(\vu^\top, \vzero^\top, -\vw^\top)^\top, (\vzero^\top, \vv^\top, -\vw^\top)^\top$ corresponding to the eigenvalue $-\lambda$.
\end{remark}



\subsection{Limiting Spectrum}

\begin{figure}[t!]
    \centering
    \input{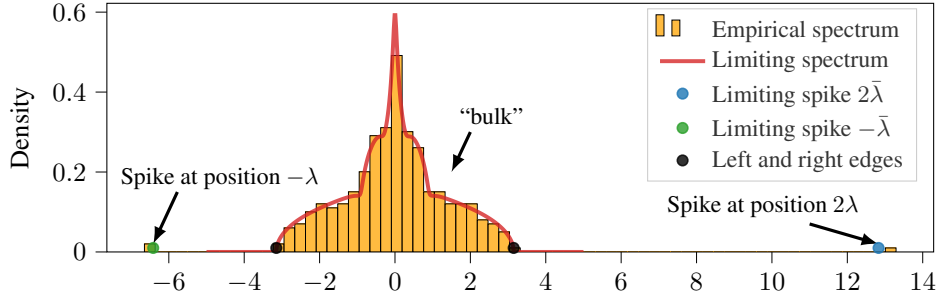}
    \vspace{-.3cm}
    \caption{Empirical versus limiting spectrum of $\mPhi$ for $\beta_T = 2, \beta_M = 3, n_1 = 130, n_2 = 80, n_3 = 140$. In addition to the ``bulk'' of eigenvalues, the spectrum of $\mPhi$ exhibits two isolated spikes at positions $2\lambda$ and $-\lambda$ as discussed in Remark \ref{remark:spikes}. In particular, the isolated spikes are accurately estimated by the limiting singular value $\bar \lambda$ as per Theorem \ref{thm:stat_unit} and Algorithm \ref{alg:summary_stats}.}
    \label{fig:spectrum_spikes}
\end{figure}

We will find subsequently that the asymptotic summary statistics $(\bar \lambda, \alpha_1, \alpha_2, \alpha_3)$ are closely related to the limiting spectral measure of the random matrix $\rmH$. Therefore, our first result characterizes precisely this limiting distribution using the Stieltjes transform formalism \citep{widder1938stieltjes}.

\begin{theorem}[Limiting spectrum]
\label{thm:spectrum_unit}
Under Assumption \ref{ass:convergence}, the empirical spectral measure of $\rmH$ or $\mPhi$ converges weakly almost surely to a deterministic distribution $\nu$ whose Stieltjes transform is given by $g(\xi) = \sum_{i=1}^3 g_i(\xi)$ such that $\Im[g(\xi)]>0$ for $\Im[\xi]>0$, and where $(g_i(\xi))_{i\in [3]}$ satisfy the following equations:
{\small\begin{align*}
        g_1(\xi) = \frac{c_1}{ g_1(\xi)  - g(\xi)  - \bar{\gamma} g_2(\xi)  - \xi },\quad
        g_2(\xi) = \frac{c_2}{ g_2(\xi)  - g(\xi)  - \bar{\gamma} g_1(\xi)  - \xi },\quad
        g_3(\xi) = \frac{c_3}{ g_3(\xi)  - g(\xi)   - \xi },
\end{align*}}
with $\bar{\gamma} = \frac{ \beta_T^2 \alpha_3^2 }{c_1 + c_2}$. In particular, the density function of $\nu$ is given by $\nu(dx) = \frac{1}{\pi} \lim_{\varepsilon\to 0} \Im \left[ g(x + i\varepsilon) \right]$.
\end{theorem}
Theorem \ref{thm:spectrum_unit} generalizes the limiting spectral measure obtained by \citep{seddik2021random} in the sense that the latter corresponds to the particular case when $\bar \gamma = 0$ (e.g. if $\beta_T = 0$).
Moreover, in the specific case $\beta_T = 0$ and $c_1 = c_2 = c_3 = \frac13$, the distribution $\nu$ describes a \textit{semi-circle law} of compact support $[-2\sqrt{2/3}, 2\sqrt{2/3}]$, and the corresponding Stieltjes transform is explicitly given by $g(\xi) = \frac34 ( - \xi + \sqrt{\xi^2 - 8/3 })$ with $g_i(\xi) = g(\xi) / 3$ for all $i\in [3]$. We refer the reader to \citep{seddik2021random} for more details and a full description of various particular cases. Moreover, an explicit formula for $g(\xi)$ can be derived in the case $c_1 = c_2$ using a formal calculation tool (e.g. SymPy). 

However, for arbitrary values of $\beta_T, \beta_M$ and of the dimension ratios $(c_1, c_2, c_3)$, the limiting spectral measure of $\rmH$ or $\mPhi$ can be computed numerically as per Algorithm \ref{alg:stieltjes_transform} which implements the equations in Theorem \ref{thm:spectrum_unit}. Figure \ref{fig:spectrum_spikes} shows that the empirical spectral measure of $\mPhi$ is accurately predicted by the limiting measure of Theorem \ref{thm:spectrum_unit} (further examples are depicted in Figure \ref{fig:specturm} in the Appendix). 
We note that the computation of $\bar \gamma$ (which is closely related to the alignment $\alpha_3$) is a key step in the numerical evaluation of $g$, which we will address next by computing the asymptotic alignments $\alpha_i$'s.

\begin{algorithm}[t!]
   \caption{Limiting Stieltjes transform as per Theorem \ref{thm:spectrum_unit}}
   \label{alg:stieltjes_transform}
\begin{algorithmic}
   \State \textbf{Input:} Complex number $\xi\in \sC\setminus \supp(\nu)$, ratios $c_1, c_2, c_3\in [0, 1]$, $\beta_T,\beta_M\geq 0$ and \texttt{option}.
   \State \textbf{Output:} Limiting Stieltjes transform $ g(\xi)$ and $ g_i(\xi)$ for $i\in [3]$.
   \State Initialize $g_1,  g_2,  g_3$ and set $g \leftarrow  g_1 +  g_2 +  g_3$.
   \If{\texttt{option} \textbf{is} \textit{``compute $\bar \gamma$''}}
   \State Compute the asymptotic summary statistics $(\bar \lambda, \alpha_1, \alpha_2, \alpha_3)$ with Algo. \ref{alg:summary_stats} and set $\bar \gamma \leftarrow \frac{\beta_T^2 \alpha_3^2}{c_1 + c_2}$.
   \EndIf
   \While{\textit{``$g$ has not converged''}}
   \If{\texttt{option} \textbf{is} \textit{``approximate $\bar \gamma$''}}
   \State Update $\bar \gamma \leftarrow \frac{\beta_T^2}{c_1 + c_2} \left( 1 - \frac{ g_3^2}{c_3} \right) $.
   \EndIf
   \State Update $ g_1 \leftarrow \frac{c_1}{ g_1 -  g - \bar \gamma  g_2 - \xi }$, $ g_2 \leftarrow \frac{c_2}{  g_2 -  g - \bar \gamma  g_1 - \xi }$, $ g_3 \leftarrow \frac{c_3}{  g_3 -  g - \xi }$, $ g \leftarrow  g_1 +  g_2 +  g_3$.
   \EndWhile
\end{algorithmic}
\end{algorithm}


\subsection{Asymptotic Summary Statistics}\label{sec:summary_stats}
In the previous subsection, we have shown that the empirical spectral measure of $\rmH$ or $\mPhi$ converges to some deterministic measure $\nu$ as we depicted in Figure \ref{fig:spectrum_spikes}. Specifically, we notice that the measure $\nu$ has a compact support that depends on the various parameters of the model. In what follows, we will need to evaluate the corresponding Stieltjes transform $g$ at the asymptotic spectral norm $\bar \lambda$, and therefore the latter must lie outside the support of $\nu$ as per the following assumption. In fact, this assumption has also been made by \citep{goulart2021random, seddik2021random}.

\begin{assumption}\label{assum:lambda_outside}
    Assume that $\bar \lambda \notin \supp(\nu)$ and $\alpha_i > 0$ for all $i\in [3]$, with $\nu$ given by Theorem \ref{thm:spectrum_unit}.
\end{assumption}

\begin{remark}[On Assumption \ref{assum:lambda_outside}]
    For any critical point $(\lambda, \vu, \vv, \vw)$ of problem (\ref{eq:deflation}), as we saw in Remark \ref{remark:spikes}, $\mPhi$ has an eigenvalue $2\lambda$. In particular, for a local maximum, $2 \lambda$ is in fact its largest eigenvalue \citep{seddik2021random}. 
    Furthermore, by studying the Hessian of that problem (which is related to $\mPhi$) at a maximum one can also show that $\lambda$ is at least as large as the second largest eigenvalue of $\mPhi$ (which is almost surely close to the right edge of the measure $\nu$). 
    Hence, the above condition in Assumption \ref{assum:lambda_outside} is slightly stronger, only requiring that inequality to hold strictly.
    See also \citep{goulart2021random} for a similar discussion in the case of a symmetric spiked tensor model.
\end{remark}

We are now in place to provide our main result which characterizes the asymptotic summary statistics $(\bar \lambda, \alpha_1, \alpha_2, \alpha_3)$ given the signal-to-noise ratios $(\beta_M, \beta_T)$ and the dimension ratios $(c_1, c_2, c_3)$.
\begin{theorem}[Asymptotic summary statistics]
\label{thm:stat_unit}
Let us define the following functions for $i\in [2]$:
\begin{small}
    \begin{align*}
    q_i(\xi) &= \sqrt{ 1 - \frac{[1 + \gamma(\xi) ] g_i^2(\xi)}{c_i}  }, \quad q_3(\xi) = \sqrt{ 1 - \frac{ g_3^2(\xi)}{c_3}  }, \quad \gamma(\xi) =  \frac{\beta_T^2 q_3^2(\xi)}{c_1 + c_2},\\
    f(\xi) &= \xi + [1 + \gamma(\xi)] g(\xi) - \gamma(\xi) g_3(\xi) - \beta_T \beta_M \prod_{i=1}^3 q_i(\xi).
\end{align*}
\end{small}
Then, under Assumptions \ref{ass:convergence} and \ref{assum:lambda_outside}, the asymptotic spectral norm $\bar \lambda$ satisfies $f(\bar{\lambda}) = 0$ and the asymptotic alignments are given by $\alpha_i = q_i(\bar \lambda)$ (in particular, $\bar \gamma = \gamma(\bar \lambda)$).
\end{theorem}

\begin{figure}[t!]
    \centering
    \input{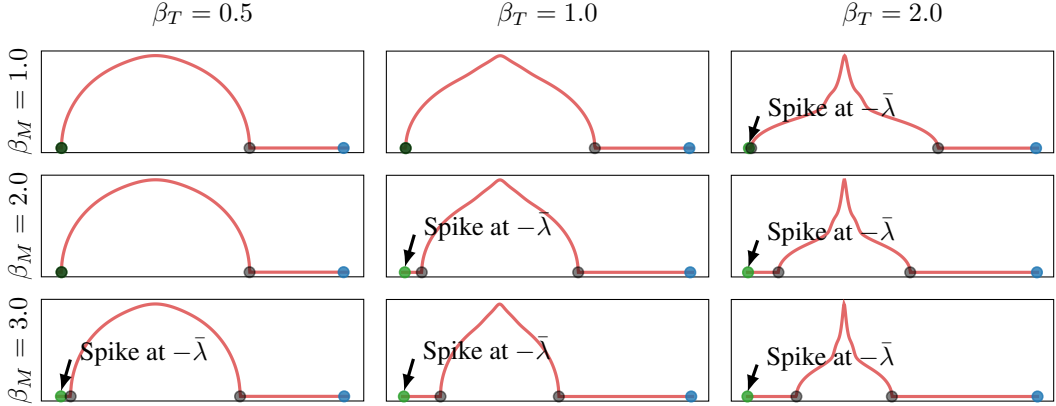}
    \vspace{-.5cm}
    \caption{Limiting spectrum and isolated spikes of $\mPhi$ for $n_1 = 80, n_2 = 100, n_3 = 90$ and varying $(\beta_T, \beta_M)$. For small values of $(\beta_T, \beta_M)$ the ``bulk'' is described by a semi-circle-like distribution. Large values of $\beta_T$ affect the ``shape'' of the limiting measure while larger values of $\beta_M$ control the position of the isolated spikes.}
    \label{fig:spectrum_spike}
\end{figure}

Theorems \ref{thm:spectrum_unit} and \ref{thm:stat_unit} show that the spectral behavior of the random matrix $\mPhi$ is fully described by its limiting spectral measure $\nu$ and the position of the limiting singular value $\bar \lambda$. This is illustrated by Figure \ref{fig:spectrum_spikes} which depicts the empirical spectrum of $\mPhi$ along with the limiting measure $\nu$ as per Theorem \ref{thm:spectrum_unit} and the asymptotic spikes computed via Theorem \ref{thm:stat_unit}. As we discussed earlier in Remark \ref{remark:spikes}, the spectrum of $\mPhi$ consists of a ``bulk'' of eigenvalues spread around $0$ and two isolated eigenvalues at positions $2\lambda$ and $-\lambda$ with multiplicities $1$ and $2$ respectively. In fact, the spike at position $-\lambda$ is only visible when the signal-to-noise ratios $(\beta_M, \beta_T)$ are large enough, and this basically corresponds to the situation where it is theoretically possible to estimate the signal components $(\vx, \vy, \vz)$ from the tensor $\tT$. In addition, note that Assumption \ref{assum:lambda_outside} holds when a spike is visible at the position $-\lambda$. This \textit{phase transition} phenomenon is highlighted in Figure \ref{fig:spectrum_spike} where we vary the signal-to-noise ratios $(\beta_M, \beta_T)$. In particular, roughly speaking, the parameter $\beta_T$ affects the ``shape'' of limiting distribution $\nu$ while $\beta_M$ determines the position of the isolated spikes. Besides, note that in the situations where $\bar\lambda$ lies inside the support of $\nu$, we solve numerically the equation $f(\bar\lambda + i\varepsilon)=0$  for some small value $\varepsilon$ (and take the real parts of $g$ and $g_i$'s), which allows us to circumvent Assumption \ref{assum:lambda_outside} in this case. 

\begin{algorithm}[t!]
   \caption{Asymptotic summary statistics as per Theorem \ref{thm:stat_unit}}
   \label{alg:summary_stats}
\begin{algorithmic}
   \State \textbf{Input:} Dimension ratios $c_1, c_2, c_3\in [0, 1]$ and signal-to-noise ratios $\beta_T, \beta_M\geq 0$.
   \State \textbf{Output:} Asymptotic summary statistics $(\bar \lambda, \alpha_1, \alpha_2, \alpha_3)$.
   \State Define $q_i(\xi) = \sqrt{ 1 - \frac{ [1 + \gamma (\xi) ] g_i^2(\xi) }{ c_i} }$ for $i\in [2]$, $q_3(\xi) = \sqrt{1 - \frac{g_3^2(\xi)}{c_3}}$ and $\gamma(\xi) = \frac{\beta_T^2 q_3^2(\xi )}{c_1 + c_2}$ where $(g_i(\xi))_{i\in [3]}$ and $g(\xi)$ are obtained by Algorithm \ref{alg:stieltjes_transform} for some $\xi \in \sR \setminus \supp(\nu)$ by setting the parameter \texttt{option} \textbf{to} \textit{``approximate $\bar \gamma$''}.
   \State Define the function $f$ as $f(\xi) = \xi + [1 + \gamma(\xi)] g(\xi) - \gamma(\xi) g_3(\xi) - \beta_T \beta_M \prod_{i=1}^3 q_i(\xi)$.
   \State Solve $f(\bar \lambda) = 0$ where the root corresponds to $\bar \lambda$ and set $\alpha_i \leftarrow q_i(\bar \lambda)$ for $i\in [3]$.
\end{algorithmic}
\end{algorithm}

\begin{figure}[t!]
    \centering
    \input{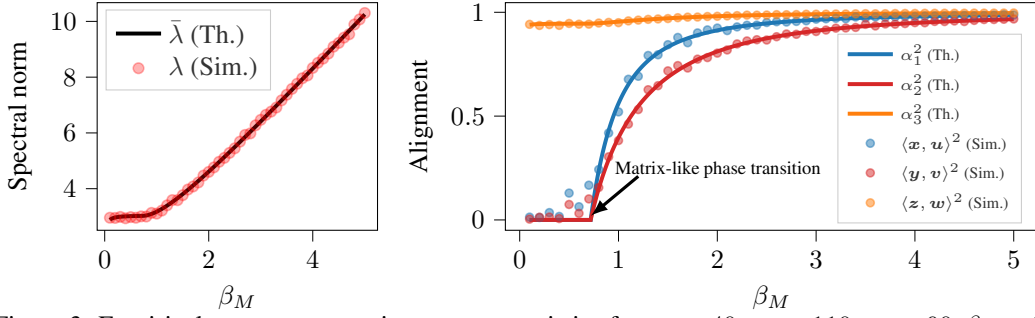}
    \vspace{-.8cm}
    \caption{Empirical versus asymptotic summary statistics for $n_1 = 40, n_2 = 110, n_3 = 90, \beta_T = 2$ and varying $\beta_M$. Simulations are obtained by averaging over $10$ independent realizations of the random matrix $\rmZ$ and tensor $\tW$. Our results exhibit a phase transition when varying $\beta_M$ above which the matrix components $(\vx, \vy)$ become estimable.}
    \label{fig:alignments}
\end{figure}

Figure \ref{fig:alignments} in turn depicts the empirical versus asymptotic summary statistics when varying the parameter $\beta_M$ (with $\beta_T$ being fixed) and shows that the empirical quantities are accurately predicted by the theoretical counterparts.
Moreover, as in standard spiked random matrix models, our results show that there exists a \textit{phase transition}, i.e., a minimum value for $\beta_M $ above which the singular vectors along the modes $1$ and $2$ $(\vu, \vv)$ start to correlate with the matrix signal components $(\vx, \vy)$. However, below this critical value of $\beta_M$, $\alpha_1$ and $\alpha_2$ are vanishing while $\alpha_3\approx 1$. The continuity of the curves of $\alpha_1$ and $\alpha_2$ when varying $\beta_M$ is a typical characteristic of spiked matrices as per the classical BBP phase transition phenomenon \citep{baik2005phase}. Besides, for smaller values of $\beta_T$ (below some critical value), the curves of $\alpha_1$ and $\alpha_2$ start to become discontinuous as per Figure \ref{fig:alignments_appendix} in Appendix \ref{appendix:simulations} which is commonly observed in spiked tensor models \citep{jagannath2020statistical}. 
In this sense, the nested matrix-tensor model is a sort of ``interpolating model'' between spiked matrices and tensors (see Appendix \ref{appendix:simulations} for additional simulations), as far as a spectral estimator of the spike is concerned.

\begin{remark}[Computation of $\alpha_3$ below the phase transition]
Even though Assumption \ref{assum:lambda_outside} is not valid in the regime where $\beta_M$ is below its critical value (because an isolated spike at position $-\bar\lambda$ outside the support of $\nu$ is not present in this case), numerically computed solutions for $f(\bar \lambda + i \varepsilon) = 0$ with a small $\varepsilon > 0$ seem to accurately estimate $\alpha_3$ as per Fig. \ref{fig:alignments} (whereas $f(\xi)$ is not defined at $\bar \lambda$ since it depends on $g(\xi)$, which is undefined inside the support of $\nu$). Yet, we currently do not have a rigorous justification for this intriguing property.
\end{remark}

\section{Application to Multi-view Clustering}

Now we illustrate the application of Theorem \ref{thm:stat_unit} to the assessment of the performance of a simple multi-view spectral clustering approach. As we presented in the introduction, we consider that we observe a tensor $\tX$ of $n$ data points of dimension $p$ along $m$ different views:
\begin{align}\label{eq:multiview}
    \tX = \left( \vmu \bar\vy^\top +  \rmZ  \right) \otimes \vh + \tW , \quad \etZ_{ij}\sim \gN\left(0, \frac{1}{p + n} \right), \, \etW_{ijk}\sim \gN \left(0, \frac{1}{p+n+m} \right),
\end{align}
where $\vmu \in \sR^p$ models the cluster means ($-\vmu$ or $\vmu$), $\bar \vy = \vy /\sqrt n$ with $\vy \in \{ -1, 1\}^n $ corresponding to the data labels (either $-1$ or $1$) and $\vh \in \sR_+^m$ is related to the variances along the different views. In particular, the case $m=1$ corresponds to the classical binary Gaussian isotropic model of centroids $\pm \vmu$ in which case the tensor $\tX$ becomes a matrix of the form $\rmX = \vmu \bar\vy^\top + \rmZ$.
\begin{wrapfigure}{r}{0.5\textwidth}
    \centering
    \input{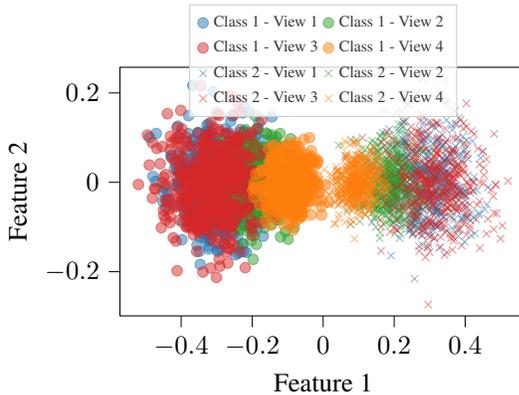}
    \vspace{-.5cm}
    \caption{Illustration of the multi-view model in Eq. (\ref{eq:multiview}) for $p=2, n=1000, m=4, \Vert \vmu \Vert = 5$ and $\Vert \vh \Vert = 3$. The first class is represented by dots and the second class by crosses. The different colors represent the views.}
    \label{fig:multiview}
\end{wrapfigure}
Figure \ref{fig:multiview} depicts the multi-view model in Eq. (\ref{eq:multiview}) for $p=2$ and $m=4$, where the first class is represented by dots and the second class is depicted by crosses, while the different views are illustrated with different colors.
Observing the tensor $\tX$, the clustering of the different data points would consist in estimating the labels vector $\vy$. Indeed, this can be performed by computing the best rank-one approximation of $\tX$ (denoted $\lambda \vu \otimes \hat \vy \otimes \vw$), and depending on the class separability condition (i.e. if $\Vert \vmu \Vert$ and $\Vert \vh\Vert$ are large enough), the $2$-mode singular vector of $\tX$ will start to correlate with $\vy$ thereby providing a clustering of the data samples. 
Our aim is to quantify the performance of this multi-view spectral clustering approach in terms of the different parameters, i.e., the dimensions $n,p,m$ and the quantities $\Vert \vmu \Vert$ and $\Vert \vh\Vert$. We precisely have the subsequent proposition which characterizes the theoretical performance of the multi-view spectral clustering method under the following growth rate assumptions.
\begin{assumption}[Growth rate]\label{assum:clustering}
    Assume that as $p, n, m \to \infty$, $\Vert \vmu \Vert, \Vert \vh \Vert = O(1)$ and denote $c_p = \lim \frac{p}{N} > 0, c_n = \lim \frac{n}{N} > 0, c_m = \lim \frac{m}{N} > 0$ with $N = p+n+m$.
\end{assumption}

\begin{proposition}[Performance of multi-view spectral clustering]\label{prop:performance_clustering} Let $\hat \vy$ be the $2^{\text{nd}}$ mode vector of the best rank-one approximation of the data tensor $\tX$. The estimated label for the sample $\tX_{:, i, j}$ is given by $\hat{\ell}_i = \sign(\hat y_i)$ for all $j\in [m]$ and let $\gL_{0/1} = \frac{1}{n}\sum_{i=1}^n \mathbb{I}\{\hat \ell_i \neq y_i\}$ be the corresponding $0/1$-loss. We have under Assumption \ref{assum:clustering}:
\begin{align*}
    (1 - \alpha^2)^{-\frac12} \left[ \sqrt{n} \hat y_i - \alpha y_i \right] \, \toind \, \gN(0, 1),
\end{align*}
where $\alpha = q_2(\bar \lambda)$ with $q_2(\cdot)$ and $\bar \lambda$ defined as per Theorem \ref{thm:stat_unit} for $(c_1, c_2, c_3) = (c_p, c_n, c_m)$ and $(\beta_M, \beta_T) = (\Vert \vmu \Vert, \Vert \vh \Vert)$. Moreover, the clustering accuracy $\max\left( \gL_{0/1}, 1 - \gL_{0/1} \right) $ converges almost surely to $\varphi\left( \alpha / \sqrt{1 - \alpha^2} \right)$ with $\varphi(x) = \frac{1}{\sqrt{2\pi}} \int_{-\infty}^x e^{-t^2/2}dt $.
\end{proposition}

\begin{figure}
    \centering
    \input{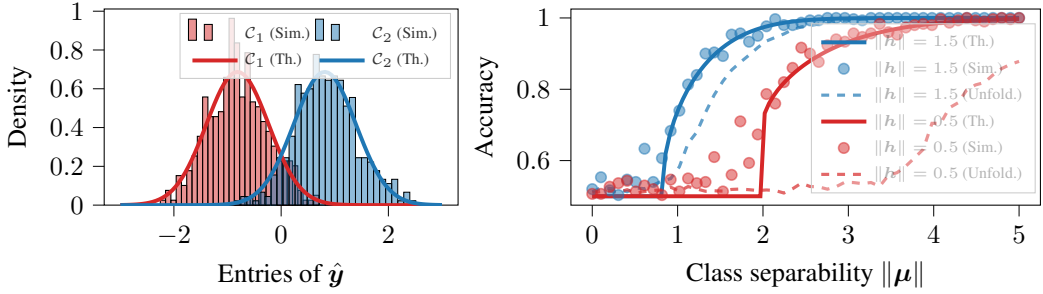}
    \vspace{-.6cm}
    \caption{\textbf{(Left)} Histogram of the entries of $\sqrt{n} \hat \vy$ for $p=200, n=800, m=100, \Vert \vmu \Vert = 1.5$ and $\Vert \vh \Vert = 2$ with the corresponding Gaussian limit as per Proposition \ref{prop:performance_clustering}. \textbf{(Right)} Empirical versus theoretical multi-view clustering performance as per Proposition \ref{prop:performance_clustering} for $p=150, n=300, m=60$ and varying $\Vert \vmu \Vert , \Vert \vh \Vert$. The dashed curves correspond to tensor unfolding which discards the rank-one structure of the data and therefore yields sub-optimal accuracy.}
    \label{fig:clustering}
\end{figure}

Proposition \ref{prop:performance_clustering} states that the entries of the 2-mode singular vector corresponding to the largest singular value of $\tX$ are Gaussian random variables, with mean and variance depending on the dimension ratios $(c_p, c_n, c_m)$ and the parameters $(\Vert \vmu \Vert, \Vert \vh \Vert)$ through the asymptotic alignment $\alpha$ obtained thanks to Theorem \ref{thm:stat_unit}. In fact, Figure \ref{fig:clustering} (Left) illustrates this Gaussianity by depicting the entries of the vector $\sqrt{n} \hat \vy$ and the corresponding normal distributions. Furthermore, the theoretical accuracy $\varphi\left( \alpha / \sqrt{1 - \alpha^2} \right)$ is also depicted in Figure \ref{fig:clustering} (Right) from which we notice that the empirical performance is accurately anticipated. Essentially, for a fixed value of $\Vert \vh \Vert$, our results show that there exists a minimal value of the class separability $\Vert \vmu \Vert$ below which the obtained accuracy is no better than a random guess, in fact, the such minimal value of $\Vert \vmu \Vert$ is related to the \textit{phase transition phenomenon} discussed in the previous section. In addition, we highlight that the considered tensor-based multi-view clustering approach provides better accuracy compared to a tensor unfolding approach, which consists in computing the top left singular vector of the unfolding of $\tX$ along the second mode \citep{arous2021long}, and therefore does not consider the hidden rank-one structure.

\section{Conclusion \& Perspectives}
We introduced the nested matrix-tensor model and provided a high-dimensional analysis of its best rank-one approximation, relying on random matrix theory. Our analysis has brought theoretical insights into the problem of muti-view clustering and demonstrates the ability of random matrix tools to assess the theoretical performance of the considered clustering method. This paves the way for an elaborated theoretical assessment and improvement of more sophisticated tensor-based methods. In particular, our present findings address only the case of binary clustering by considering the rank-one matrix model $\vmu \vy^\top + \rmZ$ which can be extended to higher ranks, thereby modeling a multi-class problem. Besides, such an extension would require the analysis of more sophisticated tensor methods (e.g. the block-term decomposition \citep{de2008decompositions}) which is more challenging compared to the present best rank-one estimator. Nevertheless, we believe our present work constitutes a fundamental basis for the development of more general results. 


\newpage
\bibliographystyle{refstyle}
\bibliography{refs}

\begin{thebibliography}{52}
\providecommand{\natexlab}[1]{#1}
\providecommand{\url}[1]{\texttt{#1}}
\expandafter\ifx\csname urlstyle\endcsname\relax
  \providecommand{\doi}[1]{doi: #1}\else
  \providecommand{\doi}{doi: \begingroup \urlstyle{rm}\Url}\fi

\bibitem[Anandkumar et~al.(2013)Anandkumar, Ge, Hsu, and
  Kakade]{AnanGHK-13-PMLR}
Anandkumar, A., Ge, R., Hsu, D., and Kakade, S.
\newblock A tensor spectral approach to learning mixed membership community
  models.
\newblock In Shalev-Shwartz, S. and Steinwart, I. (eds.), \emph{Proceedings of
  the 26th Annual Conference on Learning Theory}, volume~30 of
  \emph{Proceedings of Machine Learning Research}, pp.\  867--881, Princeton,
  NJ, USA, June 2013.

\bibitem[Anandkumar et~al.(2014)Anandkumar, Ge, Hsu, Kakade, and
  Telgarsky]{anandkumar2014tensor}
Anandkumar, A., Ge, R., Hsu, D., Kakade, S.~M., and Telgarsky, M.
\newblock Tensor decompositions for learning latent variable models.
\newblock \emph{Journal of machine learning research}, 15:\penalty0 2773--2832,
  2014.

\bibitem[Anandkumar et~al.(2015)Anandkumar, Ge, and Janzamin]{AnanGJ-15-PMLR}
Anandkumar, A., Ge, R., and Janzamin, M.
\newblock Learning overcomplete latent variable models through tensor methods.
\newblock In Gr\"unwald, P., Hazan, E., and Kale, S. (eds.), \emph{Proceedings
  of The 28th Conference on Learning Theory}, volume~40 of \emph{Proceedings of
  Machine Learning Research}, pp.\  36--112, Paris, France, 03--06 Jul 2015.

\bibitem[Auddy \& Yuan(2022)Auddy and Yuan]{auddy2022estimating}
Auddy, A. and Yuan, M.
\newblock On estimating rank-one spiked tensors in the presence of heavy tailed
  errors.
\newblock \emph{IEEE Transactions on Information Theory}, 2022.

\bibitem[Baik et~al.(2005)Baik, Ben~Arous, and P{\'e}ch{\'e}]{baik2005phase}
Baik, J., Ben~Arous, G., and P{\'e}ch{\'e}, S.
\newblock Phase transition of the largest eigenvalue for nonnull complex sample
  covariance matrices.
\newblock 2005.

\bibitem[Bakshi et~al.(2022)Bakshi, Diakonikolas, Jia, Kane, Kothari, and
  Vempala]{BaksDJK-22-STC}
Bakshi, A., Diakonikolas, I., Jia, H., Kane, D.~M., Kothari, P.~K., and
  Vempala, S.~S.
\newblock {Robustly learning mixtures of $k$ arbitrary Gaussians}.
\newblock In \emph{Proceedings of the 54th Annual ACM Symposium on Theory of
  Computing}, pp.\  1234--1247, Rome, Italy, June 2022.

\bibitem[Ben~Arous et~al.(2021)Ben~Arous, Huang, and Huang]{arous2021long}
Ben~Arous, G., Huang, D.~Z., and Huang, J.
\newblock Long random matrices and tensor unfolding.
\newblock \emph{arXiv preprint arXiv:2110.10210}, 2021.

\bibitem[Ben~Arous et~al.(2022)Ben~Arous, Gheissari, and
  Jagannath]{ben2022high}
Ben~Arous, G., Gheissari, R., and Jagannath, A.
\newblock High-dimensional limit theorems for sgd: Effective dynamics and
  critical scaling.
\newblock \emph{Advances in Neural Information Processing Systems},
  35:\penalty0 25349--25362, 2022.

\bibitem[Blum \& Mitchell(1998)Blum and Mitchell]{blum1998combining}
Blum, A. and Mitchell, T.
\newblock Combining labeled and unlabeled data with co-training.
\newblock In \emph{Proceedings of the eleventh annual conference on
  Computational learning theory}, pp.\  92--100, 1998.

\bibitem[Cao et~al.(2015)Cao, Zhang, Fu, Liu, and Zhang]{cao2015diversity}
Cao, X., Zhang, C., Fu, H., Liu, S., and Zhang, H.
\newblock Diversity-induced multi-view subspace clustering.
\newblock In \emph{Proceedings of the IEEE conference on computer vision and
  pattern recognition}, pp.\  586--594, 2015.

\bibitem[Chaudhuri et~al.(2009)Chaudhuri, Kakade, Livescu, and
  Sridharan]{chaudhuri2009multi}
Chaudhuri, K., Kakade, S.~M., Livescu, K., and Sridharan, K.
\newblock Multi-view clustering via canonical correlation analysis.
\newblock In \emph{Proceedings of the 26th annual international conference on
  machine learning}, pp.\  129--136, 2009.

\bibitem[Chen et~al.(2021)Chen, Handschy, and Lerman]{chen2021phase}
Chen, W.-K., Handschy, M., and Lerman, G.
\newblock Phase transition in random tensors with multiple independent spikes.
\newblock \emph{The Annals of Applied Probability}, 31\penalty0 (4):\penalty0
  1868--1913, 2021.

\bibitem[Couillet \& Benaych-Georges(2016)Couillet and
  Benaych-Georges]{couillet2016kernel}
Couillet, R. and Benaych-Georges, F.
\newblock Kernel spectral clustering of large dimensional data.
\newblock 2016.

\bibitem[Dasgupta et~al.(2001)Dasgupta, Littman, and
  McAllester]{dasgupta2001pac}
Dasgupta, S., Littman, M., and McAllester, D.
\newblock Pac generalization bounds for co-training.
\newblock \emph{Advances in neural information processing systems}, 14, 2001.

\bibitem[De~Lathauwer(2008)]{de2008decompositions}
De~Lathauwer, L.
\newblock Decompositions of a higher-order tensor in block terms—part ii:
  Definitions and uniqueness.
\newblock \emph{SIAM Journal on Matrix Analysis and Applications}, 30\penalty0
  (3):\penalty0 1033--1066, 2008.

\bibitem[Fernandes et~al.(2021)Fernandes, Fanaee-T, and Gama]{FernFG-21-AIR}
Fernandes, S., Fanaee-T, H., and Gama, J.
\newblock Tensor decomposition for analysing time-evolving social networks: An
  overview.
\newblock \emph{Artificial Intelligence Review}, 54:\penalty0 2891--2916, 2021.

\bibitem[Gao et~al.(2015)Gao, Nie, Li, and Huang]{gao2015multi}
Gao, H., Nie, F., Li, X., and Huang, H.
\newblock Multi-view subspace clustering.
\newblock In \emph{Proceedings of the IEEE international conference on computer
  vision}, pp.\  4238--4246, 2015.

\bibitem[Ge et~al.(2015)Ge, Huang, and Kakade]{GeHK-15-STC}
Ge, R., Huang, Q., and Kakade, S.~M.
\newblock Learning mixtures of {Gaussians} in high dimensions.
\newblock In \emph{Proceedings of the 47th annual ACM Symposium on Theory of
  Computing}, pp.\  761--770, Portland, OR, USA, June 2015.

\bibitem[Goulart et~al.(2022)Goulart, Couillet, and Comon]{goulart2021random}
Goulart, J. H.~{\relax de M}., Couillet, R., and Comon, P.
\newblock A random matrix perspective on random tensors.
\newblock \emph{Journal on Machine Learning Research}, 23\penalty0
  (264):\penalty0 1--36, 2022.

\bibitem[Gujral et~al.(2020)Gujral, Pasricha, and Papalexakis]{GujrPP-20-TWC}
Gujral, E., Pasricha, R., and Papalexakis, E.
\newblock Beyond rank-1: Discovering rich community structure in multi-aspect
  graphs.
\newblock In \emph{Proceedings of The Web Conference 2020}, pp.\  452--462,
  Taipei, Taiwan, April 2020.

\bibitem[Hachem et~al.(2007)Hachem, Loubaton, and
  Najim]{hachem2007deterministic}
Hachem, W., Loubaton, P., and Najim, J.
\newblock Deterministic equivalents for certain functionals of large random
  matrices.
\newblock \emph{The Annals of Applied Probability}, 17\penalty0 (3):\penalty0
  875--930, 2007.

\bibitem[Hsu \& Kakade(2013)Hsu and Kakade]{HsuK-13-CITCS}
Hsu, D. and Kakade, S.~M.
\newblock Learning mixtures of spherical {Gaussians}: moment methods and
  spectral decompositions.
\newblock In \emph{Proceedings of the 4th conference on Innovations in
  Theoretical Computer Science}, pp.\  11--20, Berkeley, CA, USA, January 2013.

\bibitem[Hsu et~al.(2012)Hsu, Kakade, and Zhang]{HsuKZ-12-JCSC}
Hsu, D., Kakade, S.~M., and Zhang, T.
\newblock A spectral algorithm for learning hidden {Markov} models.
\newblock \emph{Journal of Computer and System Sciences}, 78\penalty0
  (5):\penalty0 1460--1480, 2012.

\bibitem[Huang et~al.(2015)Huang, Niranjan, H., and Anandkumar]{HuanNH-15-JMLR}
Huang, F., Niranjan, U.~N., H., M.~U., and Anandkumar, A.
\newblock Online tensor methods for learning latent variable models.
\newblock \emph{Journal of Machine Learning Research}, 16:\penalty0 2797--2835,
  2015.

\bibitem[Jagannath et~al.(2020)Jagannath, Lopatto, and
  Miolane]{jagannath2020statistical}
Jagannath, A., Lopatto, P., and Miolane, L.
\newblock Statistical thresholds for tensor {PCA}.
\newblock \emph{The Annals of Applied Probability}, 30\penalty0 (4):\penalty0
  1910--1933, 2020.

\bibitem[Janzamin et~al.(2019)Janzamin, Ge, Kossaifi, and
  Anandkumar]{JanzGKA-19-FTML}
Janzamin, M., Ge, R., Kossaifi, J., and Anandkumar, A.
\newblock Spectral learning on matrices and tensors.
\newblock \emph{Foundations and Trends in Machine Learning}, 12\penalty0
  (5-6):\penalty0 393--536, 2019.

\bibitem[Kakade \& Foster(2007)Kakade and Foster]{kakade2007multi}
Kakade, S.~M. and Foster, D.~P.
\newblock Multi-view regression via canonical correlation analysis.
\newblock In \emph{Learning Theory: 20th Annual Conference on Learning Theory,
  COLT 2007, San Diego, CA, USA; June 13-15, 2007. Proceedings 20}, pp.\
  82--96. Springer, 2007.

\bibitem[Kargas \& Sidiropoulos(2019)Kargas and Sidiropoulos]{KargS-19-PMLR}
Kargas, N. and Sidiropoulos, N.~D.
\newblock Learning mixtures of smooth product distributions: Identifiability
  and algorithm.
\newblock In Chaudhuri, K. and Sugiyama, M. (eds.), \emph{Proceedings of the
  22nd International Conference on Artificial Intelligence and Statistics},
  volume~89 of \emph{Proceedings of Machine Learning Research}, pp.\  388--396,
  Naha, Okinawa, Japan, Apr 2019.

\bibitem[Khouja et~al.(2022)Khouja, Mattei, and Mourrain]{KhouMM-22-JSC}
Khouja, R., Mattei, P.~A., and Mourrain, B.
\newblock {Tensor decomposition for learning Gaussian mixtures from moments}.
\newblock \emph{Journal of Symbolic Computation}, 113:\penalty0 193--210, 2022.

\bibitem[Kilmer et~al.(2013)Kilmer, Braman, Hao, and Hoover]{kilmer2013third}
Kilmer, M.~E., Braman, K., Hao, N., and Hoover, R.~C.
\newblock Third-order tensors as operators on matrices: A theoretical and
  computational framework with applications in imaging.
\newblock \emph{SIAM Journal on Matrix Analysis and Applications}, 34\penalty0
  (1):\penalty0 148--172, 2013.

\bibitem[Kofidis \& Regalia(2002)Kofidis and Regalia]{KofiR-02-SIMAX}
Kofidis, E. and Regalia, P.~A.
\newblock On the best rank-1 approximation of higher-order supersymmetric
  tensors.
\newblock \emph{SIAM Journal on Matrix Analysis and Applications}, 23\penalty0
  (3):\penalty0 863--884, 2002.

\bibitem[Lesieur et~al.(2017)Lesieur, Miolane, Lelarge, Krzakala, and
  Zdeborov{\'a}]{lesieur2017statistical}
Lesieur, T., Miolane, L., Lelarge, M., Krzakala, F., and Zdeborov{\'a}, L.
\newblock Statistical and computational phase transitions in spiked tensor
  estimation.
\newblock In \emph{2017 IEEE International Symposium on Information Theory
  (ISIT)}, pp.\  511--515. IEEE, 2017.

\bibitem[Lim(2005)]{lim2005singular}
Lim, L.-H.
\newblock Singular values and eigenvalues of tensors: a variational approach.
\newblock In \emph{Proc. IEEE International Workshop on Computational Advances
  in Multi-Sensor Adaptive Processing (CAMSAP)}, pp.\  129--132, 2005.

\bibitem[Liu et~al.(2013)Liu, Ji, Glänzel, and De~Moor]{6193101}
Liu, X., Ji, S., Glänzel, W., and De~Moor, B.
\newblock Multiview partitioning via tensor methods.
\newblock \emph{IEEE Transactions on Knowledge and Data Engineering},
  25\penalty0 (5):\penalty0 1056--1069, 2013.
\newblock \doi{10.1109/TKDE.2012.95}.

\bibitem[Nigam \& Ghani(2000)Nigam and Ghani]{nigam2000analyzing}
Nigam, K. and Ghani, R.
\newblock Analyzing the effectiveness and applicability of co-training.
\newblock In \emph{Proceedings of the ninth international conference on
  Information and knowledge management}, pp.\  86--93, 2000.

\bibitem[Oseledets \& Kharyuk(2021)Oseledets and Kharyuk]{OselK-21-CMMP}
Oseledets, I.~V. and Kharyuk, P.~V.
\newblock Structuring data with block term decomposition: Decomposition of
  joint tensors and variational block term decomposition as a parametrized
  mixture distribution model.
\newblock \emph{Computational Mathematics and Mathematical Physics},
  61\penalty0 (5):\penalty0 816--835, 2021.

\bibitem[Papalexakis et~al.(2012)Papalexakis, Sidiropoulos, and
  Bro]{PapaSB-12-TSP}
Papalexakis, E.~E., Sidiropoulos, N.~D., and Bro, R.
\newblock From k-means to higher-way co-clustering: Multilinear decomposition
  with sparse latent factors.
\newblock \emph{IEEE Transactions on Signal Processing}, 61\penalty0
  (2):\penalty0 493--506, 2012.

\bibitem[Perry et~al.(2020)Perry, Wein, and Bandeira]{perry2016statistical}
Perry, A., Wein, A.~S., and Bandeira, A.~S.
\newblock Statistical limits of spiked tensor models.
\newblock In \emph{Annales de l'Institut Henri Poincar{\'e}, Probabilit{\'e}s
  et Statistiques}, volume~56, pp.\  230--264. Institut Henri Poincar{\'e},
  2020.

\bibitem[Rahmani et~al.(2020)Rahmani, Niranjan, Fay, Takeda, and
  Brodzki]{RahmNFT-20-PRL}
Rahmani, D., Niranjan, M., Fay, D., Takeda, A., and Brodzki, J.
\newblock {Estimation of Gaussian mixture models via tensor moments with
  application to online learning}.
\newblock \emph{Pattern Recognition Letters}, 131:\penalty0 285--292, 2020.

\bibitem[Richard \& Montanari(2014)Richard and
  Montanari]{montanari2014statistical}
Richard, E. and Montanari, A.
\newblock A statistical model for tensor {PCA}.
\newblock \emph{Advances in neural information processing systems}, 27, 2014.

\bibitem[Seddik et~al.(2021)Seddik, Guillaud, and Couillet]{seddik2021random}
Seddik, M. E.~A., Guillaud, M., and Couillet, R.
\newblock When random tensors meet random matrices.
\newblock \emph{arXiv preprint arXiv:2112.12348}, 2021.

\bibitem[Singhal et~al.(2023)Singhal, Mirza, Rajwade, and
  Gurumoorthy]{SingMRG-23-arxiv}
Singhal, P., Mirza, W., Rajwade, A., and Gurumoorthy, K.~S.
\newblock Estimating joint probability distribution with low-rank tensor
  decomposition, {Radon} transforms and dictionaries.
\newblock \emph{arXiv:2304.08740}, 2023.

\bibitem[Stein(1981)]{stein1981estimation}
Stein, C.~M.
\newblock Estimation of the mean of a multivariate normal distribution.
\newblock \emph{The annals of Statistics}, pp.\  1135--1151, 1981.

\bibitem[Sun(2013)]{sun2013survey}
Sun, S.
\newblock A survey of multi-view machine learning.
\newblock \emph{Neural computing and applications}, 23:\penalty0 2031--2038,
  2013.

\bibitem[Tao(2012)]{tao2012topics}
Tao, T.
\newblock \emph{Topics in random matrix theory}, volume 132.
\newblock American Mathematical Soc., 2012.

\bibitem[Van Der~Vaart \& Wellner(1996)Van Der~Vaart and Wellner]{van1996weak}
Van Der~Vaart, A.~W. and Wellner, J.~A.
\newblock Weak convergence.
\newblock In \emph{Weak convergence and empirical processes}, pp.\  16--28.
  Springer, 1996.

\bibitem[White et~al.(2012)White, Zhang, Schuurmans, and Yu]{white2012convex}
White, M., Zhang, X., Schuurmans, D., and Yu, Y.-l.
\newblock Convex multi-view subspace learning.
\newblock \emph{Advances in neural information processing systems}, 25, 2012.

\bibitem[Widder(1938)]{widder1938stieltjes}
Widder, D.~V.
\newblock The stieltjes transform.
\newblock \emph{Transactions of the American Mathematical Society}, 43\penalty0
  (1):\penalty0 7--60, 1938.

\bibitem[Wu et~al.(2020)Wu, Xie, Nie, Lin, and Zha]{wu2020unified}
Wu, J., Xie, X., Nie, L., Lin, Z., and Zha, H.
\newblock Unified graph and low-rank tensor learning for multi-view clustering.
\newblock In \emph{Proceedings of the AAAI conference on artificial
  intelligence}, volume~34, pp.\  6388--6395, 2020.

\bibitem[Xie et~al.(2018)Xie, Tao, Zhang, Liu, Zhang, and Qu]{xie2018unifying}
Xie, Y., Tao, D., Zhang, W., Liu, Y., Zhang, L., and Qu, Y.
\newblock On unifying multi-view self-representations for clustering by tensor
  multi-rank minimization.
\newblock \emph{International Journal of Computer Vision}, 126:\penalty0
  1157--1179, 2018.

\bibitem[Xu et~al.(2013)Xu, Tao, and Xu]{xu2013survey}
Xu, C., Tao, D., and Xu, C.
\newblock A survey on multi-view learning.
\newblock \emph{arXiv preprint arXiv:1304.5634}, 2013.

\bibitem[Zhao et~al.(2017)Zhao, Xie, Xu, and Sun]{zhao2017multi}
Zhao, J., Xie, X., Xu, X., and Sun, S.
\newblock Multi-view learning overview: Recent progress and new challenges.
\newblock \emph{Information Fusion}, 38:\penalty0 43--54, 2017.

\end{thebibliography}

\newpage
\appendix

\begin{center}
    \begin{Large}
    \textbf{Supplementary Material: A Nested Matrix-Tensor Model for Noisy Multi-view Clustering}
    \end{Large}
\end{center}

\section{Technical Proofs}\label{appendix:proofs}

Throughout this appendix, we consider the same statistical model as in the main paper but with general variance $\sigma_T^2$ (resp. $\sigma_M^2$) for the noise tensor (resp. matrix):
\begin{equation}
\label{eq:model}
    \tT = \beta_T \rmM \otimes \vz + \frac{1}{\sqrt{n_T}} \tW \, \in \mathbb{R}^{n_1 \times n_2 \times n_3}, \quad
    \rmM = \beta_M \vx \otimes \vy + \frac{1}{\sqrt{n_M}} \rmZ \, \in \mathbb{R}^{n_1 \times n_2 },
\end{equation}
where we assume that $\|\vx\|=\|\vy\|=\|\vz\|=1$ and that the entries of $\tW$ and $\rmZ$ are independent Gaussian random variables, i.e., $\etW_{ijk} \sim \gN(0, \sigma_T^2)$ and $\etZ_{ij} \sim \gN(0, \sigma_M^2)$.

We now state (and then prove) the following results for general variances extending \Cref{thm:spectrum_unit,thm:stat_unit}
beyond the standard case $\sigma_T^2=\sigma_M^2=1$.

\begin{theorem}[Limiting spectrum for general variances]
\label{thm:spectrum_gen}
Under Assumption \ref{ass:convergence}, the empirical spectral measure of $\rmH$ converges weakly almost surely to a deterministic distribution $\mu$ whose Stieltjes transform is given by $g(\xi) = \sum_{i=1}^3 g_i(\xi)$ such that $\Im[g(\xi)]>0$ for $\Im[\xi]>0$, and where $(g_i(\xi))_{i\in [3]}$ satisfy the following equations
{\small\begin{align*}
        g_1(\xi) &= \frac{c_1}{ \sigma_T^2 ( g_1(\xi)  - g(\xi) )  - \bar{\gamma} \sigma_M^2 g_2(\xi)  - \xi },\quad
        g_2(\xi) = \frac{c_2}{ \sigma_T^2 ( g_2(\xi)  - g(\xi) )  - \bar{\gamma} \sigma_M^2 g_1(\xi)  - \xi },\\ 
        \quad g_3(\xi) &= \frac{c_3}{ \sigma_T^2 ( g_3(\xi)  - g(\xi) )   - \xi }
\end{align*}}
with $\bar{\gamma} = \frac{ \beta_T^2 \alpha_3^2 }{c_1 + c_2}$.
\end{theorem}

\begin{theorem}[Asymptotic summary statistics for general variances]
\label{thm:stat_gen}
Let us define the following functions for $i\in [2]$:
\begin{small}
    \begin{align*}
    q_i(\xi) &= \sqrt{ 1 - \frac{[\sigma_T^2 + \sigma_M^2 \gamma(\xi) ] g_i^2(\xi)}{c_i}  }, \quad q_3(\xi) = \sqrt{ 1 - \frac{ \sigma_T^2 g_3^2(\xi)}{c_3}  }, \quad \gamma(\xi) =  \frac{\beta_T^2 q_3^2(\xi)}{c_1 + c_2},\\
    f(\xi) &= \xi + [\sigma_T^2 + \sigma_M^2 \gamma(\xi)] g(\xi) - \sigma_M^2 \gamma(\xi) g_3(\xi) - \beta_T \beta_M \prod_{i=1}^3 q_i(\xi).
\end{align*}
\end{small}
Then, under Assumptions \ref{ass:convergence} and \ref{assum:lambda_outside}, the asymptotic spectral norm $\bar \lambda$ satisfies $f(\bar{\lambda}) = 0$ and the asymptotic alignments are given by $\alpha_i = q_i(\bar \lambda)$ (in particular, $\bar \gamma = \gamma(\bar \lambda)$).
\end{theorem}

The proofs are provided below: they heavily rely on a classical RMT identity known as Stein's lemma (a.k.a. Gaussian integration by parts) that we recall next.

\begin{lemma}[Stein's lemma \citep{stein1981estimation}]
    Let $X \sim \mathcal{N}(0, \sigma^2)$ and $f$ a continuously differentiable function having at most polynomial growth, then
    \begin{equation*}
        \mathbb{E}[X f(X)] = \sigma^2 \mathbb{E}[f'(X)] \ ,
    \end{equation*}
    given that the above expectations exist.
\end{lemma}

\paragraph{Additional notations.}
Without loss of generality (up to replacing $\vx$ by $-\vx$ or $\vu$ by $-\vu$), let us assume that $\langle \vu, \vx \rangle \to \alpha_1$ and (for similar reason) say $\langle \vv, \vy \rangle \to \alpha_2$ and $\langle \vw, \vz \rangle \to \alpha_3$.
To maintain consistency, we will adhere to the convention of using $\evx_i$ or $\evu_i$ to represent the components of the first mode, $\evy_j$ or $\evv_j$ to represent the components of the second mode and $\evz_k$ or $\evw_k$ to represent the components of the third mode. In the remainder, if some quantity expresses as $a(n) = \sum_{i=1}^r b_i(n)$, the notation $a(n)\asymp b_j(n)$ for some $j\in [r]$ means that $b_j(n)$ is the only contributing term of $a(n)$ for estimating the quantities of interest (limiting spectrum, summary statistics) as $n\to \infty$:
this approximation is considered based on the results developed in \citep{seddik2021random} which identified such contributing terms.

\subsection{Proof of \Cref{thm:spectrum_gen}}
We start by expressing the expected value of $\lambda$ via Eq. (\ref{eq:eigen}):

\begin{multline}
\label{eq:expect_lambda}
    \mathbb{E}[\lambda] = \mathbb{E}[ \tT( \vu, \vv, \vw ) ]
    = \mathbb{E}[ \beta_T (\rmM \otimes \vz)( \vu, \vv, \vw ) ] + \frac{1}{\sqrt{n_T}} \mathbb{E}[ \tW( \vu, \vv, \vw ) ] \\
    = \underbrace{ \mathbb{E}[ \beta_T \beta_M \langle \vx, \vu \rangle \langle \vy, \vv \rangle \langle \vz, \vw \rangle ] }_{ \to \beta_T \beta_M \alpha_1 \alpha_2 \alpha_3 } + \beta_T \mathbb{E}[ \frac{1}{\sqrt{n_M}} \vu^\top \rmZ \vv \underbrace{ \langle \vz, \vw \rangle }_{ \to \alpha_3 } ] + \frac{1}{\sqrt{n_T}} \mathbb{E}[ \tW( \vu, \vv, \vw ) ] \ .
\end{multline}

Then, let us rewrite the last term in Eq. (\ref{eq:expect_lambda}) by using Stein's lemma:
\begin{multline*}
    \frac{1}{\sqrt{n_T}} \mathbb{E}[ \tW( \vu, \vv, \vw ) ]
    = \frac{1}{\sqrt{n_T}} \sum_{ijk} \mathbb{E}[ \evu_i \evv_j \evw_k \etW_{ijk} ]
    = \frac{1}{\sqrt{n_T}} \sum_{ijk} \sigma_T^2 \mathbb{E}[ \frac{\partial ( \evu_i \evv_j \evw_k )}{ \partial \etW_{ijk} } ] \\
    = \frac{1}{\sqrt{n_T}} \sum_{ijk} \sigma_T^2 \mathbb{E}[ \frac{\partial \evu_i }{ \partial \etW_{ijk} } \evv_j \evw_k ]
    + \frac{1}{\sqrt{n_T}} \sum_{ijk} \sigma_T^2 \mathbb{E}[ \evu_i \frac{\partial \evv_j }{ \partial \etW_{ijk} } \evw_k ]
    + \frac{1}{\sqrt{n_T}} \sum_{ijk} \sigma_T^2 \mathbb{E}[ \evu_i \evv_j \frac{\partial \evw_k }{ \partial \etW_{ijk} } ] \ .
\end{multline*}

In order to compute the partial derivatives appearing in the equation above, we recall from Eq. (\ref{eq:eigen}) that
\begin{equation*}
    \tT(\cdot, \vv, \vw) = \lambda \vu
    \iff \forall i, \quad 
    \lambda \evu_i = \sum_{jk} \evv_j \evw_k \etT_{ijk} \ ,
\end{equation*}
which implies:
\begin{multline*}
    \frac{\partial \lambda}{\partial \etW_{abc}} \evu_i + \lambda \frac{\partial \evu_i}{\partial \etW_{abc}} 
    = \sum_{jk} \frac{\partial \evv_j}{\partial \etW_{abc}} \evw_k \etT_{ijk}
    + \sum_{jk} \evv_j \frac{\partial \evw_k}{\partial \etW_{abc}} \etT_{ijk}
    + \sum_{jk} \evv_j \evw_k \underbrace{ \frac{\partial \etT_{ijk} }{\partial \etW_{abc}} }_{ \frac{1}{\sqrt{n_T}} \delta_{ia} \delta_{jb} \delta_{kc} } \\
    = [ \tT(\vw) \frac{\partial \vv}{\partial \etW_{abc}} ]_i + [ \tT(\vv) \frac{\partial \vw}{\partial \etW_{abc}} ]_i
    + \frac{1}{\sqrt{n_T}} \evv_b \evw_c \delta_{ia} \ .
\end{multline*}
Equivalently, in a more compact vectorial form,
\begin{equation}
\label{eq:dudW}
    \frac{\partial \lambda}{\partial \etW_{abc}} \vu + \lambda \frac{\partial \vu}{\partial \etW_{abc}} 
    = \tT(\vw) \frac{\partial \vv}{\partial \etW_{abc}} + \tT(\vv) \frac{\partial \vw}{\partial \etW_{abc}} 
    + \frac{1}{\sqrt{n_T}} \evv_b \evw_c \ve_{a}^{n_1} \ ,
\end{equation}
where $\ve_{a}^{n_1}$ denotes the standard basis vector of dimension $n_1$ with all coordinates equal to 0 except at position $a$, where it is equal to 1.

Similarly, one can show that
\begin{equation}
\label{eq:dvdW}
    \frac{\partial \lambda}{\partial \etW_{abc}} \vv + \lambda \frac{\partial \vv}{\partial \etW_{abc}} 
    = \tT(\vw)^\top \frac{\partial \vu}{\partial \etW_{abc}} + \tT(\vu) \frac{\partial \vw}{\partial \etW_{abc}} 
    + \frac{1}{\sqrt{n_T}} \evu_a \evw_c \ve_{b}^{n_2} \ ,
\end{equation}
and
\begin{equation}
\label{eq:dwdW}
    \frac{\partial \lambda}{\partial \etW_{abc}} \vw + \lambda \frac{\partial \vw}{\partial \etW_{abc}}
    = \tT(\vu)^\top \frac{\partial \vv}{\partial \etW_{abc}} + \tT(\vv)^\top \frac{\partial \vu}{\partial \etW_{abc}} 
    + \frac{1}{\sqrt{n_T}} \evu_a \evv_b \ve_{c}^{n_3} \ .
\end{equation}

Let us now compute $\frac{\partial \lambda}{\partial \etW_{abc}}$. We have by Eq. (\ref{eq:eigen}),
\begin{equation*}
    \lambda = \tT(\vu, \vv, \vw) = \sum_{ijk} \evu_i \evv_j \evw_k \etT_{ijk} \ ,
\end{equation*}
from which we deduce:
\begin{multline}
\label{eq:sumzero}
    \frac{\partial \lambda}{\partial \etW_{abc}}
    = \sum_{ijk} \frac{ \partial \evu_i }{ \partial \etW_{abc} } \evv_j \evw_k \etT_{ijk}
    + \sum_{ijk} \evu_i \frac{ \partial \evv_j }{ \partial \etW_{abc} } \evw_k \etT_{ijk}
    + \sum_{ijk} \evu_i \evv_j \frac{ \partial \evw_k }{ \partial \etW_{abc} } \etT_{ijk}
    + \sum_{ijk} \frac{\evu_i \evv_j \evw_k}{\sqrt{n_T}} \delta_{ia} \delta_{jb} \delta_{kc} \\
    = \frac{\evu_a \evv_b \evw_c}{\sqrt{n_T}} \ ,
\end{multline}
where the last equality follows from observing that the three first sums in Eq. (\ref{eq:sumzero}) are equal to zero.
Indeed, recalling that $\|\vu\|^2$ is constant equal to $1$, we have for the first sum:
\begin{equation}
    \sum_{ijk} \frac{ \partial \evu_i }{ \partial \etW_{abc} } \evv_j \evw_k \etT_{ijk}
    = \tT( \frac{\partial \vu}{ \partial \etW_{abc} } , \vv, \vw)
    = \lambda \vu^\top \frac{\partial \vu}{ \partial \etW_{abc} } 
    = \frac{\lambda}{2} \frac{\partial \|\vu\|^2  }{ \partial \etW_{abc} } 
    = 0 \ ,
\end{equation}
and the same holds for the two other sums.

By combining \Cref{eq:dudW,eq:dvdW,eq:dwdW,eq:sumzero}, we have
\begin{equation}
    \begin{bmatrix}
    \frac{\partial \vu}{\partial \etW_{abc}} \\
    \frac{\partial \vv}{\partial \etW_{abc}} \\
    \frac{\partial \vw}{\partial \etW_{abc}}
    \end{bmatrix}
    = 
    -\frac{1}{\sqrt{n_T}}
    \underbrace{
    \left(
    \underbrace{
    \begin{bmatrix}
    0 & \tT(\vw) & \tT(\vv) \\
    \tT(\vw)^\top & 0 & \tT(\vu) \\
    \tT(\vv)^\top & \tT(\vu)^\top & 0
    \end{bmatrix}
    }_{\mPhi}
    - \lambda \mI_{n_T}
    \right)^{-1}
    }_{ R(\lambda) }
    \begin{bmatrix}
    \evv_b \evw_c (\ve_{a}^{n_1} - \evu_a \vu) \\
    \evu_a \evw_c (\ve_{b}^{n_2} - \evv_b \vv) \\
    \evu_a \evv_b (\ve_{c}^{n_3} - \evw_c \vw)
    \end{bmatrix} \ .
\end{equation}

Similarly, one can also show that

\begin{equation}
    \begin{bmatrix}
    \frac{\partial \vu}{\partial \ermZ_{ab}} \\
    \frac{\partial \vv}{\partial \ermZ_{ab}} \\
    \frac{\partial \vw}{\partial \ermZ_{ab}}
    \end{bmatrix}
    = 
    -\frac{\beta_T}{\sqrt{n_M}}
    \left(
    \mPhi
    - \lambda \mI_{n_T}
    \right)^{-1}
    \begin{bmatrix}
    \langle \vw, \vz \rangle \evv_b (\ve_{a}^{n_1} - \evu_a \vu) \\
    \langle \vw, \vz \rangle \evu_a (\ve_{b}^{n_2} - \evv_b \vv) \\
    \evu_a \evv_b \vz - \langle \vw, \vz \rangle \evu_a \evv_b \vw
    \end{bmatrix} \ .
\end{equation}

By discarding asymptotically negligible terms in $\mPhi$, we consider the asymptotically equivalent matrix $\rmH\asymp \mPhi= \rmH + \rmL$ given by:
\begin{equation}
    \rmH = \begin{bmatrix}
    0 & \frac{ \langle \vw, \vz \rangle \beta_T }{ \sqrt{n_M} } \rmZ + \frac{1}{\sqrt{n_T}} \tW(\vw) & \frac{1}{\sqrt{n_T}} \tW(\vv) \\
    \frac{ \langle \vw, \vz \rangle \beta_T }{ \sqrt{n_M} } \rmZ^\top + \frac{1}{\sqrt{n_T}} \tW(\vw)^\top & 0 & \frac{1}{\sqrt{n_T}} \tW(\vu) \\
    \frac{1}{\sqrt{n_T}} \tW(\vv)^\top & \frac{1}{\sqrt{n_T}} \tW(\vu)^\top & 0
    \end{bmatrix} \ ,
\end{equation}
such that $\rmL = \mPhi - \rmH $ is equal to
\begin{small}
    \begin{equation}\label{eq:expression_of_L}
    \begin{bmatrix}
    0 & \langle \vw, \vz \rangle \beta_T \beta_M  \vx \otimes \vy & \langle \vv, \vy \rangle \beta_T \beta_M  \vx \otimes \vz + \frac{\beta_T}{\sqrt{n_M}} (\rmZ \vy) \otimes \vz \\
    \langle \vw, \vz \rangle \beta_T \beta_M  \vy \otimes \vx & 0 & \langle \vu, \vx \rangle \beta_T \beta_M  \vy \otimes \vz + \frac{\beta_T}{\sqrt{n_M}} (\rmZ^\top \vx) \otimes \vz \\
    \langle \vv, \vy \rangle \beta_T \beta_M  \vz \otimes \vx + \frac{\beta_T}{\sqrt{n_M}} \vz \otimes (\rmZ \vy) & \langle \vu, \vx \rangle \beta_T \beta_M  \vz \otimes \vy + \frac{\beta_T}{\sqrt{n_M}} \vz \otimes (\rmZ^\top \vx) & 0
    \end{bmatrix} \ .
\end{equation}
\end{small}

For convenience, let us denote $ \beta_T' = \langle \vw, \vz \rangle \beta_T $
and the matrix-valued function $\rmG$:
\begin{equation}
    \rmG(\xi) = ( \rmH - \xi \mI_{n_T} )^{-1}
    = \begin{bmatrix}
    \rmG^{11}(\xi) & \rmG^{12}(\xi) & \rmG^{13}(\xi) \\
    \rmG^{12}(\xi)^{\top} & \rmG^{22}(\xi) & \rmG^{23}(\xi) \\
    \rmG^{13}(\xi)^{\top} & \rmG^{23}(\xi)^\top & \rmG^{33}(\xi)
    \end{bmatrix} \ ,
\end{equation}
which implies that $ \rmH \rmG(\xi) = \xi \rmG(\xi) + \mI_{n_T} $.

Plus, observe that
\begin{align}
    \label{eq:partial_deriv_W}
    \frac{\partial \vu}{\partial \etW_{abc}} & \asymp - \frac{1}{\sqrt{n_T}} \rmG^{11} \evv_b \evw_c \ve_{a}^{n_1}
    \quad \Rightarrow \quad \frac{\partial \evu_a}{\partial \etW_{abc}} \asymp - \frac{1}{\sqrt{n_T}} \ermG^{11}_{aa} \evv_b \evw_c \ , \\
    \frac{\partial \vv}{\partial \etW_{abc}} & \asymp - \frac{1}{\sqrt{n_T}} \rmG^{22} \evu_a \evw_c \ve_{b}^{n_2}
    \quad \Rightarrow \quad \frac{\partial \evv_b}{\partial \etW_{abc}} \asymp - \frac{1}{\sqrt{n_T}} \ermG^{22}_{bb} \evu_a \evw_c \ , \\
    \frac{\partial \vw}{\partial \etW_{abc}} & \asymp - \frac{1}{\sqrt{n_T}} \rmG^{33} \evu_a \evv_b \ve_{c}^{n_3}
    \quad \Rightarrow \quad \frac{\partial \evw_c}{\partial \etW_{abc}} \asymp - \frac{1}{\sqrt{n_T}} \ermG^{33}_{cc} \evu_a \evv_b \ ,
\end{align}
and
\begin{align}
    \label{eq:partial_deriv_Z}
    \frac{\partial \vu}{\partial \ermZ_{ab}} & \asymp - \frac{\beta_T}{\sqrt{n_M}} \alpha_3 \evv_b \rmG^{11} \ve_{a}^{n_1}
    \quad \Rightarrow \quad \frac{\partial \evu_a}{\partial \ermZ_{ab}} \asymp - \frac{\beta_T}{\sqrt{n_M}} \ermG^{11}_{aa} \alpha_3 \evv_b \ , \\
    \frac{\partial \vv}{\partial \ermZ_{ab}} & \asymp - \frac{\beta_T}{\sqrt{n_M}} \alpha_3 \evu_a \rmG^{22} \ve_{b}^{n_2}
    \quad \Rightarrow \quad \frac{\partial \evv_b}{\partial \ermZ_{ab}} \asymp - \frac{\beta_T}{\sqrt{n_M}} \ermG^{22}_{bb} \alpha_3 \evu_a \ .
\end{align}

Let us now compute $ \frac{1}{n_T} \mathbb{E}[ \Tr( \rmG ) ] $.
First note that $\Tr( \rmH \rmG )= \Tr( \rmH \rmG )^{11} + \Tr( \rmH \rmG )^{22} + \Tr( \rmH \rmG )^{33}$
decomposes as the sum of the traces of three blocks.

\paragraph{Block (1,1)}
We have
\begin{equation}
    ( \rmH \rmG )^{11} = \left[ \frac{\beta_T'}{\sqrt{n_M}} \rmZ + \frac{1}{\sqrt{n_T}} \tW(\vw) \right] \rmG^{12 \top} + \frac{1}{\sqrt{n_T}} \tW(\vv) \rmG^{13 \top}
\end{equation}
and thus
\begin{multline*}
    ( \rmH \rmG )^{11}_{ii} = \left( \left[ \frac{\beta_T'}{\sqrt{n_M}} \rmZ + \frac{1}{\sqrt{n_T}} \tW(\vw) \right] \rmG^{12 \top} \right)_{ii} + \frac{1}{\sqrt{n_T}} \left( \tW(\vv) \rmG^{13 \top} \right)_{ii} \\
    = \left( \frac{\beta_T'}{\sqrt{n_M}} \rmZ \rmG^{12 \top} \right)_{ii}  + \frac{1}{\sqrt{n_T}} 
    \left( \tW(\vw) \rmG^{12 \top} \right)_{ii} + \frac{1}{\sqrt{n_T}} \sum_{jk} \evv_{j} \etW_{ijk} \ermG_{ik}^{13} \\
    = \frac{\beta_T'}{\sqrt{n_M}} \sum_{j} \ermZ_{ij} \ermG_{ij}^{12}  + \frac{1}{\sqrt{n_T}} 
    \sum_{jk} \evw_{k} \etW_{ijk} \ermG_{ij}^{12} + \frac{1}{\sqrt{n_T}} \sum_{jk} \evv_{j} \etW_{ijk} \ermG_{ik}^{13} .
\end{multline*}
We deduce that
\begin{multline}
    \frac{1}{n_T} \mathbb{E}[ \Tr( \rmH \rmG )^{11} ]
    = \frac{1}{n_T} \sum_{i} \mathbb{E}[ ( \rmH \rmG )^{11}_{ii} ] \\
    = \frac{\beta_T'}{n_T \sqrt{n_M}} \sum_{ij} \mathbb{E}[ \ermZ_{ij} \ermG_{ij}^{12} ]  + \frac{1}{n_T \sqrt{n_T}} \sum_{ijk} \mathbb{E}[ \evw_{k} \etW_{ijk} \ermG_{ij}^{12} ] + \frac{1}{n_T \sqrt{n_T}} \sum_{ijk} \mathbb{E}[ \evv_{j} \etW_{ijk} \ermG_{ik}^{13} ] \ .
\end{multline}
Once again, we will use Stein's lemma to compute $\mathbb{E}[ \ermZ_{ij} \ermG_{ij}^{12} ]$. For that purpose,
we need the derivative of $\rmG$ w.r.t. $\ermZ_{ij}$:
\begin{multline*}
    \rmH \rmG = \xi \rmG + \mI_{n_T} 
    \Rightarrow 
    \frac{\partial \rmH }{ \partial \ermZ_{ij} } \rmG
    + \rmH \frac{\partial \rmG }{ \partial \ermZ_{ij} }
    = \xi \frac{\partial \rmG }{ \partial \ermZ_{ij} } 
    \Rightarrow 
    ( \rmH - \xi \mI_{n_T} ) \frac{\partial \rmG }{ \partial \ermZ_{ij} }
    = - \frac{\partial \rmH }{ \partial \ermZ_{ij} } \rmG \\
    \Rightarrow 
    \frac{\partial \rmG }{ \partial \ermZ_{ij} }
    = - \rmG \frac{\partial \rmH }{ \partial \ermZ_{ij} } \rmG .
\end{multline*}

In the last equation, observe that
\begin{multline*}
    \frac{\partial \rmH }{ \partial \ermZ_{ij} } =
    \begin{bmatrix}
    0 & \frac{ \beta_T \langle \vw, \vz \rangle }{ \sqrt{n_M} } \ve_{i}^{n_1} \ve_{j}^{n_2 \top} + \langle \vz, \frac{\partial \vw}{\partial \ermZ_{ij}} \rangle \frac{\beta_T}{\sqrt{n_M}} \rmZ & 0 \\
    \frac{ \beta_T \langle \vw, \vz \rangle }{ \sqrt{n_M} } \ve_{j}^{n_2} \ve_{i}^{n_1 \top} + \langle \vz, \frac{\partial \vw}{\partial \ermZ_{ij}} \rangle \frac{\beta_T}{\sqrt{n_M}} \rmZ^\top & 0 & 0 \\
    0 & 0 & 0
    \end{bmatrix} \\
    \asymp
    \begin{bmatrix}
    0 & \frac{ \beta_T \langle \vw, \vz \rangle }{ \sqrt{n_M} } \ve_{i}^{n_1} \ve_{j}^{n_2 \top} & 0 \\
    \frac{ \beta_T \langle \vw, \vz \rangle }{ \sqrt{n_M} } \ve_{j}^{n_2} \ve_{i}^{n_1 \top} & 0 & 0 \\
    0 & 0 & 0
    \end{bmatrix} \ .
\end{multline*}

Hence,
\begin{equation*}
    \frac{\partial \rmG }{ \partial \ermZ_{ij} } \asymp
    - \frac{ \beta_T' }{ \sqrt{n_M} } \begin{bmatrix}
    \rmG^{12} \ve_{j}^{n_2} \ve_{i}^{n_1 \top} & \rmG^{11} \ve_{i}^{n_1} \ve_{j}^{n_2 \top} & 0 \\
    \rmG^{22} \ve_{j}^{n_2} \ve_{i}^{n_1 \top} & \rmG^{12 \top} \ve_{i}^{n_1} \ve_{j}^{n_2 \top} & 0 \\
    \rmG^{23 \top} \ve_{j}^{n_2} \ve_{i}^{n_1 \top} & \rmG^{13 \top} \ve_{i}^{n_1} \ve_{j}^{n_2 \top} & 0
    \end{bmatrix} \rmG \ ,
\end{equation*}
from which it follows that
\begin{multline}
    \frac{\partial \ermG^{12}_{ij} }{ \partial \ermZ_{ij} }
    \asymp 
    - \frac{ \beta_T' }{ \sqrt{n_M} } [ \rmG^{12} \ve_{j}^{n_2} \ve_{i}^{n_1 \top} \rmG^{12} ]_{ij}
    - \frac{ \beta_T' }{ \sqrt{n_M} } [ \rmG^{11} \ve_{i}^{n_1} \ve_{j}^{n_2 \top} \rmG^{22} ]_{ij}
    = - \frac{ \beta_T' }{ \sqrt{n_M} } \ermG^{12}_{ij} \ermG^{12}_{ij}
    - \frac{ \beta_T' }{ \sqrt{n_M} } \ermG^{11}_{ii} \ermG^{22}_{jj} \\
    \asymp - \frac{ \beta_T' }{ \sqrt{n_M} } \ermG^{11}_{ii} \ermG^{22}_{jj} \ ,
\end{multline}

and finally by Stein's lemma,
\begin{multline}
    \frac{\beta_T'}{n_T \sqrt{n_M}} \sum_{ij} \mathbb{E}[ \ermZ_{ij} \ermG_{ij}^{12} ]
    = \frac{\beta_T'}{n_T \sqrt{n_M}} \sigma_M^2 \sum_{ij} \mathbb{E}\left[ \frac{\partial \ermG^{12}_{ij} }{ \partial \ermZ_{ij} } \right] 
    \asymp - \frac{ \beta_T'^2 }{ n_T n_M } \sigma_M^2 \sum_{ij} \mathbb{E}[ \ermG^{11}_{ii} \ermG^{22}_{jj} ] \\
    = - \beta_T'^2 \left(\frac{n_T}{n_M} \right) \sigma_M^2 \mathbb{E}\left[ \underbrace{ \frac{1}{n_T} \Tr(\rmG^{11}) }_{ \rightarrow g_1(\xi) } \underbrace{ \frac{1}{n_T} \Tr(\rmG^{22}) }_{ \rightarrow g_2(\xi) } \right] \ .
\end{multline}
For ease of notation throughout this proof, we will often omit the argument $\xi$ and simply write $g_i$ instead of $g_i(\xi)$.

Similarly, we can show that
\begin{equation}
    \frac{\partial \ermG^{12}_{ij} }{ \partial \etW_{ijk} }
    \asymp - \frac{ 1 }{ \sqrt{n_T} } \evw_k \ermG^{11}_{ii} \ermG^{22}_{jj} \ ,
\end{equation}
and 
\begin{equation}
    \frac{\partial \ermG^{13}_{ik} }{ \partial \etW_{ijk} }
    \asymp - \frac{ 1 }{ \sqrt{n_T} } \evv_j \ermG^{11}_{ii} \ermG^{33}_{kk} \ .
\end{equation}

Hence,
\begin{multline}
    \frac{1}{n_T} \mathbb{E}[ \Tr( \rmH \rmG )^{11} ]
    \asymp - \frac{\beta_T'^2 n_T }{ n_M } \sigma_M^2 g_1 g_2
    + \frac{\sigma_T^2}{n_T \sqrt{n_T}} \sum_{ijk} \mathbb{E}[ \evw_{k} \frac{\partial \ermG_{ij}^{12} }{\partial \etW_{ijk} } ]
    + \frac{\sigma_T^2}{n_T \sqrt{n_T}} \sum_{ijk} \mathbb{E}[ \evv_{j} \frac{\partial \ermG_{ik}^{13} }{\partial \etW_{ijk} } ] \\
    \asymp - \frac{\beta_T'^2 n_T }{ n_M } \sigma_M^2 g_1 g_2
    - \frac{\sigma_T^2}{n_T^2} \sum_{ijk} \mathbb{E}[ \evw_{k}^2 \ermG^{11}_{ii} \ermG^{22}_{jj} ]
    - \frac{\sigma_T^2}{n_T^2} \sum_{ijk} \mathbb{E}[ \evv_{j}^2 \ermG^{11}_{ii} \ermG^{33}_{kk} ] \\
    \asymp - \left( \sigma_T^2 + \frac{\beta_T'^2 n_T }{ n_M } \sigma_M^2 \right) g_1 g_2 - \sigma_T^2 g_1 g_3 \ ,
\end{multline}
where $g_3(\xi)$ denotes the limit of $\frac{1}{n_T} \mathbb{E}[\Tr(\rmG^{33})]$.

\paragraph{Block (2,2)}
We have
\begin{equation}
    ( \rmH \rmG )^{22} = \left[ \frac{\beta_T'}{\sqrt{n_M}} \rmZ^\top + \frac{1}{\sqrt{n_T}} \tW(\vw)^\top \right] \rmG^{12} + \frac{1}{\sqrt{n_T}} \tW(\vu) \rmG^{23 \top}
\end{equation}
and thus
\begin{equation*}
    ( \rmH \rmG )^{22}_{jj} 
    = \frac{\beta_T'}{\sqrt{n_M}} \sum_{i} \ermZ_{ij} \ermG_{ij}^{12}  + \frac{1}{\sqrt{n_T}} 
    \sum_{ik} \evw_{k} \etW_{ijk} \ermG_{ij}^{12} + \frac{1}{\sqrt{n_T}} \sum_{ik} \evu_{i} \etW_{ijk} \ermG_{jk}^{23} .
\end{equation*}
We deduce that
\begin{multline}
    \frac{1}{n_T} \mathbb{E}[ \Tr( \rmH \rmG )^{22} ]
    = \frac{1}{n_T} \sum_{j} \mathbb{E}[ ( \rmH \rmG )^{22}_{jj} ] \\
    = \frac{\beta_T'}{n_T \sqrt{n_M}} \sum_{ij} \mathbb{E}[ \ermZ_{ij} \ermG_{ij}^{12} ]  + \frac{1}{n_T \sqrt{n_T}} \sum_{ijk} \mathbb{E}[ \evw_{k} \etW_{ijk} \ermG_{ij}^{12} ] + \frac{1}{n_T \sqrt{n_T}} \sum_{ijk} \mathbb{E}[ \evu_{i} \etW_{ijk} \ermG_{jk}^{23} ] \ .
\end{multline}

Then, we show that
\begin{equation}
    \frac{\partial \ermG^{23}_{jk} }{ \partial \etW_{ijk} }
    \asymp - \frac{ 1 }{ \sqrt{n_T} } \evu_i \ermG^{22}_{jj} \ermG^{33}_{kk} \ .
\end{equation}

Hence,
\begin{equation}
    \frac{1}{n_T} \mathbb{E}[ \Tr( \rmH \rmG )^{22} ]
    \asymp - \left( \sigma_T^2 + \frac{\beta_T'^2 n_T }{ n_M } \sigma_M^2 \right) g_1 g_2 - \sigma_T^2 g_2 g_3 \ .
\end{equation}

\paragraph{Block (3,3)}
We have
\begin{equation}
    ( \rmH \rmG )^{33} = \frac{1}{\sqrt{n_T}} \tW(\vv)^\top \rmG^{13} + \frac{1}{\sqrt{n_T}} \tW(\vu)^\top \rmG^{23}
\end{equation}
and thus
\begin{equation*}
    ( \rmH \rmG )^{33}_{kk} 
    = \frac{1}{\sqrt{n_T}} \sum_{ij} \evv_{j} \etW_{ijk} \ermG_{ik}^{13} + \frac{1}{\sqrt{n_T}} \sum_{ij} \evu_{i} \etW_{ijk} \ermG_{jk}^{23} .
\end{equation*}
We deduce that
\begin{multline}
    \frac{1}{n_T} \mathbb{E}[ \Tr( \rmH \rmG )^{33} ]
    = \frac{1}{n_T} \sum_{k} \mathbb{E}[ ( \rmH \rmG )^{33}_{kk} ] \\
    = \frac{1}{n_T \sqrt{n_T}} \sum_{ijk} \mathbb{E}[ \evv_{j} \etW_{ijk} \ermG_{ik}^{13} ] + \frac{1}{n_T \sqrt{n_T}} \sum_{ijk} \mathbb{E}[ \evu_{i} \etW_{ijk} \ermG_{jk}^{23} ] \ .
\end{multline}

Hence,
\begin{equation}
    \frac{1}{n_T} \mathbb{E}[ \Tr( \rmH \rmG )^{33} ]
    \asymp - \sigma_T^2 g_1 g_3 - \sigma_T^2 g_2 g_3 \ .
\end{equation}

\paragraph{System of equations for traces.}
Observe that
\begin{align}
    \frac{1}{n_T} \mathbb{E}[ \Tr ( \rmH \rmG )^{11} ] 
    & = \xi \frac{1}{n_T} \mathbb{E}[ \Tr( \rmG^{11} ) ] + \underbrace{ \frac{n_1}{n_T} }_{ \to c_1 } \ , \\
    \frac{1}{n_T} \mathbb{E}[ \Tr ( \rmH \rmG )^{22} ] 
    & = \xi \frac{1}{n_T} \mathbb{E}[ \Tr( \rmG^{22} ) ] + \underbrace{ \frac{n_2}{n_T} }_{ \to c_2 } \ ,\\
    \frac{1}{n_T} \mathbb{E}[ \Tr ( \rmH \rmG )^{33} ] 
    & = \xi \frac{1}{n_T} \mathbb{E}[ \Tr( \rmG^{33} ) ] + \underbrace{ \frac{n_3}{n_T} }_{ \to c_3 } \ .
\end{align}

Hence, the block-wise traces $g_1(\xi),g_2(\xi),g_3(\xi)$ satisfy the following system of equations:
\begin{equation}
\begin{cases}
    g_1 = \frac{- c_1 }{ \sigma_T^2 (g - g_1) + \bar{\gamma} \sigma_M^2 g_2 + \xi } \\
    g_2 = \frac{- c_2 }{ \sigma_T^2 (g - g_2) + \bar{\gamma} \sigma_M^2 g_1 + \xi } \\
    g_3 = \frac{- c_3 }{ \sigma_T^2 (g - g_3) + \xi } \\
    g = g_1 + g_2 + g_3
\end{cases} \ .
\end{equation}

\subsection{Proof of \Cref{thm:stat_gen}}

Throughout the proof for notational convenience, the traces $g_i, g$ are evaluated at $\bar{\lambda}$ whenever their argument $\xi$ is not specified.

\paragraph{First alignment}

Let us start with the first alignment $\langle \vu, \vx \rangle$.

\begin{multline}
    \mathbb{E}[ \lambda \langle \vu, \vx \rangle ] = \mathbb{E}[ \tT( \vx, \vv, \vw ) ]
    = \mathbb{E}[ \beta_T (\rmM \otimes \vz)( \vx, \vv, \vw ) ] + \frac{1}{\sqrt{n_T}} \mathbb{E}[ \tW( \vx, \vv, \vw ) ] \\
    = \underbrace{ \mathbb{E}[ \beta_T \beta_M \langle \vy, \vv \rangle \langle \vz, \vw \rangle ] }_{ \to \beta_T \beta_M \alpha_2 \alpha_3 } + \beta_T \mathbb{E}[ \frac{1}{\sqrt{n_M}} \vx^\top \rmZ \vv \underbrace{ \langle \vz, \vw \rangle }_{ \to \alpha_3 } ] + \frac{1}{\sqrt{n_T}} \mathbb{E}[ \tW( \vx, \vv, \vw ) ] \ .
\end{multline}

Then, let us rewrite the last term by using Stein's lemma:
\begin{multline}
    \frac{1}{\sqrt{n_T}} \mathbb{E}[ \tW( \vx, \vv, \vw ) ]
    = \frac{1}{\sqrt{n_T}} \sum_{ijk} \evx_i \mathbb{E}[ \evv_j \evw_k \etW_{ijk} ]
    = \frac{1}{\sqrt{n_T}} \sum_{ijk} \evx_i \sigma_T^2 \mathbb{E}[ \frac{\partial ( \evv_j \evw_k )}{ \partial \etW_{ijk} } ] \\
    = \frac{1}{\sqrt{n_T}} \sum_{ijk} \evx_i \sigma_T^2 \mathbb{E}[ \frac{\partial \evv_j }{ \partial \etW_{ijk} } \evw_k ]
    + \frac{1}{\sqrt{n_T}} \sum_{ijk} \evx_i \sigma_T^2 \mathbb{E}[ \evv_j \frac{\partial \evw_k }{ \partial \etW_{ijk} } ] \ .
\end{multline}

By using Eq. (\ref{eq:partial_deriv_W}),
\begin{multline}
    \frac{1}{\sqrt{n_T}} \mathbb{E}[ \tW( \vx, \vv, \vw ) ]
    \asymp - \frac{1}{n_T} \sum_{ijk} \evx_i \sigma_T^2 \mathbb{E}[ \ermG^{22}_{jj} \evu_i \evw_k^2 ]
    - \frac{1}{n_T} \sum_{ijk} \evx_i \sigma_T^2 \mathbb{E}[ \ermG^{33}_{kk} \evu_i \evv_j^2 ] \\
    \asymp - \sigma_T^2 g_2 \mathbb{E}[ \langle \vx, \vu \rangle ] - \sigma_T^2 g_3 \mathbb{E}[ \langle \vx, \vu \rangle ]
    = - \sigma_T^2 \mathbb{E}[ \langle \vx, \vu \rangle ] ( g_2 + g_3 ) \ .
\end{multline}

Similarly, we have
\begin{equation}
    \frac{\beta_T \alpha_3}{\sqrt{n_M}} \mathbb{E}[ \vx^\top \rmZ \vv ]
    = \frac{\beta_T \alpha_3}{\sqrt{n_M}} \sum_{ij} \evx_i \mathbb{E}[ \evv_j \ermZ_{ij} ]
    = \frac{\beta_T \alpha_3}{\sqrt{n_M}} \sum_{ij} \evx_i \sigma_M^2 \mathbb{E}[ \frac{\partial \evv_j}{\partial \ermZ_{ij}} ] \ ,
\end{equation}
where the partial derivative is given by Eq. (\ref{eq:partial_deriv_Z}).
Hence,
\begin{equation}
    \frac{\beta_T \alpha_3}{\sqrt{n_M}} \mathbb{E}[ \vx^\top \rmZ \vv ]
    \asymp - \frac{\beta_T^2 \alpha_3^2 n_T }{n_M} \sigma_M^2 \mathbb{E}[ \langle \vx, \vu \rangle ] g_2 \ .
\end{equation}

Finally,
\begin{multline}
    \mathbb{E}[ \lambda \langle \vu, \vx \rangle ]
    \asymp \beta_T \beta_M \alpha_2 \alpha_3 + \frac{\beta_T \alpha_3}{\sqrt{n_M}} \mathbb{E}[ \vx^\top \rmZ \vv ] + \frac{1}{\sqrt{n_T}} \mathbb{E}[ \tW( \vx, \vv, \vw ) ] \\
    \asymp \beta_T \beta_M \alpha_2 \alpha_3 - \frac{\beta_T^2 \alpha_3^2 }{c_1 + c_2} \sigma_M^2 \mathbb{E}[ \langle \vx, \vu \rangle ] g_2 - \sigma_T^2 \mathbb{E}[ \langle \vx, \vu \rangle ] ( g_2 + g_3 ) \\
    \quad \Rightarrow \quad \alpha_1 = 
    \frac{ \beta_T \beta_M \alpha_2 \alpha_3 }{ \bar{\lambda} + \bar{\gamma} \sigma_M^2 g_2 + \sigma_T^2 ( g - g_1 ) } \ .
\end{multline}

We can further simplify the expression of this alignment by using Eq. (\ref{eq:lambda}):
\begin{equation}
    \alpha_1^2 = 
    \frac{ \beta_T \beta_M \alpha_1 \alpha_2 \alpha_3 }{ \bar{\lambda} + \bar{\gamma} \sigma_M^2 g_2 + \sigma_T^2 ( g - g_1 ) }
    =
    \frac{ \bar{\lambda} + \bar{\gamma} \sigma_M^2 (g_1+g_2) + \sigma_T^2 g }{ \bar{\lambda} + \bar{\gamma} \sigma_M^2 g_2 + \sigma_T^2 ( g - g_1 ) } \ ,
\end{equation}
which implies that
\begin{equation}
    \label{eq:alpha_x}
    \boxed{
    \alpha_1 = \sqrt{ \frac{ \bar{\lambda} + \bar{\gamma} \sigma_M^2 (g_1+g_2) + \sigma_T^2 g }{ \bar{\lambda} + \bar{\gamma} \sigma_M^2 g_2 + \sigma_T^2 ( g - g_1 ) } }
    } \ .
\end{equation}

\paragraph{Second alignment}
By symmetry with the first alignment,
\begin{equation}
    \label{eq:alpha_y}
    \boxed{
    \alpha_2 = \sqrt{ \frac{ \bar{\lambda} + \bar{\gamma} \sigma_M^2 (g_1+g_2) + \sigma_T^2 g }{ \bar{\lambda} + \bar{\gamma} \sigma_M^2 g_1 + \sigma_T^2 ( g - g_2 ) } }
    } \ .
\end{equation}

\paragraph{Third alignment}

Let us compute the third and last alignment $\langle \vw, \vz \rangle$.

\begin{multline}
    \mathbb{E}[ \lambda \langle \vw, \vz \rangle ] = \mathbb{E}[ \tT( \vu, \vv, \vz ) ]
    = \mathbb{E}[ \beta_T (\rmM \otimes \vz)( \vu, \vv, \vz ) ] + \frac{1}{\sqrt{n_T}} \mathbb{E}[ \tW( \vu, \vv, \vz ) ] \\
    = \underbrace{ \mathbb{E}[ \beta_T \beta_M \langle \vx, \vu \rangle \langle \vy, \vv \rangle ] }_{ \to \beta_T \beta_M \alpha_1 \alpha_2 } + \beta_T \mathbb{E}[ \frac{1}{\sqrt{n_M}} \vu^\top \rmZ \vv ] + \frac{1}{\sqrt{n_T}} \mathbb{E}[ \tW( \vu, \vv, \vz ) ] \ .
\end{multline}

Then, let us rewrite the last term by using Stein's lemma:
\begin{multline}
    \frac{1}{\sqrt{n_T}} \mathbb{E}[ \tW( \vu, \vv, \vz ) ]
    = \frac{1}{\sqrt{n_T}} \sum_{ijk} \evz_k \mathbb{E}[ \evu_i \evv_j \etW_{ijk} ]
    = \frac{1}{\sqrt{n_T}} \sum_{ijk} \evz_k \sigma_T^2 \mathbb{E}[ \frac{\partial ( \evu_i \evv_j )}{ \partial \etW_{ijk} } ] \\
    = \frac{1}{\sqrt{n_T}} \sum_{ijk} \evz_k \sigma_T^2 \mathbb{E}[ \frac{\partial \evu_i }{ \partial \etW_{ijk} } \evv_j ]
    + \frac{1}{\sqrt{n_T}} \sum_{ijk} \evz_k \sigma_T^2 \mathbb{E}[ \evu_i \frac{\partial \evv_j }{ \partial \etW_{ijk} } ] \ .
\end{multline}

By using Eq. (\ref{eq:partial_deriv_W}),
\begin{align}
    \frac{1}{\sqrt{n_T}} \mathbb{E}[ \tW( \vu, \vv, \vz ) ]
    & \asymp - \frac{1}{n_T} \sum_{ijk} \evz_k \sigma_T^2 \mathbb{E}[ \ermG^{11}_{ii} \evv_j^2 \evw_k ]
    - \frac{1}{n_T} \sum_{ijk} \evz_k \sigma_T^2 \mathbb{E}[ \ermG^{22}_{jj} \evu_i^2 \evw_k ] \\
    & \asymp - \sigma_T^2 \alpha_3 ( g_1 + g_2 ) \ .
\end{align}

Further, it holds from Eq. (\ref{eq:partial_deriv_Z}),
\begin{multline}
    \frac{\beta_T}{\sqrt{n_M}} \mathbb{E}[ \vu^\top \rmZ \vv ]
    = \frac{\beta_T}{\sqrt{n_M}} \sum_{ij} \mathbb{E}[ \evu_i \evv_j \ermZ_{ij} ]
    = \frac{\beta_T}{\sqrt{n_M}} \sigma_M^2 \sum_{ij} \mathbb{E}[ \frac{\partial \evu_i}{ \partial \ermZ_{ij}} \evv_j ]
    + \frac{\beta_T}{\sqrt{n_M}} \sigma_M^2 \sum_{ij} \mathbb{E}[ \evu_i \frac{\partial \evv_j}{ \partial \ermZ_{ij}} ] \\
    \asymp - \frac{\beta_T^2 \alpha_3 n_T}{ n_M } \sigma_M^2 ( g_1 + g_2 ) 
    = - \frac{\bar{\gamma} \sigma_M^2}{ \alpha_3 } ( g_1 + g_2 ) \ .
\end{multline}

By combining the previous formulas, we obtain:
\begin{align}
    \bar{\lambda} \alpha_3 & =
    \beta_T \beta_M \alpha_1 \alpha_2 - \sigma_T^2 \alpha_3 ( g_1 + g_2 ) - \frac{\bar{\gamma} \sigma_M^2}{ \alpha_3 } ( g_1 + g_2 ) \\
    \quad \Rightarrow \quad
    \bar{\lambda} \alpha_3^2 & = \beta_T \beta_M \alpha_1 \alpha_2 \alpha_3 - \sigma_T^2 \alpha_3^2 ( g_1 + g_2 ) - \bar{\gamma} \sigma_M^2 ( g_1 + g_2 ) \\
    \quad \Rightarrow \quad
    \alpha_3^2 & = \frac{\bar{\lambda} + \sigma_T^2 g }{ \bar{\lambda} + \sigma_T^2( g - g_3 ) } \ ,
\end{align}
which implies that
\begin{equation}
    \label{eq:alpha_z}
    \boxed{
    \alpha_3 = \sqrt{ \frac{\bar{\lambda} + \sigma_T^2 g }{ \bar{\lambda} + \sigma_T^2( g - g_3 ) } }
    } \ .
\end{equation}

\paragraph{Simplification:} By the definition of $(g_1, g_2, g_3)$, we have
\begin{align*}
    \bar{\lambda} + \bar{\gamma} \sigma_M^2 g_2 + \sigma_T^2 ( g - g_1 ) = \frac{-c_1}{g_1}, \quad \bar{\lambda} + \bar{\gamma} \sigma_M^2 g_1 + \sigma_T^2 ( g - g_2 ) = \frac{-c_2}{g_2}, \quad \bar{\lambda} + \sigma_T^2 ( g - g_3 ) = \frac{-c_3}{g_3}
\end{align*}
Therefore (for $ \alpha_1$),
\begin{multline*}
    \alpha_1^2 = \frac{ \bar{\lambda} + \bar{\gamma} \sigma_M^2 (g_1 + g_2) + \sigma_T^2 g }{ \bar{\lambda} + \bar{\gamma} \sigma_M^2 g_2 + \sigma_T^2 ( g - g_1 ) } = \frac{ \bar{\lambda} + \bar{\gamma} \sigma_M^2 g_2 + \sigma_T^2 ( g - g_1 )  + (\sigma_T^2 + \bar{\gamma} \sigma_M^2) g_1 }{ \bar{\lambda} + \bar{\gamma} \sigma_M^2 g_2 + \sigma_T^2 ( g - g_1 ) } \\ 
    = 1 + \frac{(\sigma_T^2+\bar{\gamma} \sigma_M^2) g_1}{\bar{\lambda} + \bar{\gamma} \sigma_M^2 g_2 + \sigma_T^2 ( g - g_1 ) } = 1 - \frac{(\sigma_T^2+\bar{\gamma} \sigma_M^2) g_1^2}{c_1} 
\end{multline*}
From this (and similarly for the other alignments), we find that
\begin{align}
    \boxed{ \alpha_1 = \sqrt{ 1 - \frac{(\sigma_T^2+\bar{\gamma} \sigma_M^2) g_1^2(\bar{\lambda})}{c_1}  }, \quad \alpha_2 = \sqrt{ 1 - \frac{(\sigma_T^2+\bar{\gamma} \sigma_M^2) g_2^2(\bar{\lambda})}{c_2}  }, \quad \alpha_3 = \sqrt{ 1 - \frac{\sigma_T^2 g_3^2(\bar{\lambda})}{c_3}  } } \ .
\end{align}

Define the following functions for $i\in [2]$
\begin{align}
    q_i(\xi) = \sqrt{ 1 - \frac{[\sigma_T^2 + \sigma_M^2 \gamma(\xi) ] g_i^2(\xi)}{c_i}  }, \quad q_3(\xi) = \sqrt{ 1 - \frac{ \sigma_T^2 g_3^2(\xi)}{c_3}  }
\end{align}
with $\gamma(\xi) =  \frac{\beta_T^2 q_3^2(\xi)}{c_1 + c_2} $.
We also find that $\bar{\lambda}$ satisfies $f(\bar{\lambda}) = 0$ with
\begin{align}
    \boxed{ f(\xi) = \xi + [\sigma_T^2 + \sigma_M^2 \gamma(\xi)] g(\xi) - \sigma_M^2 \gamma(\xi) g_3(\xi) - \beta_T \beta_M \prod_{i=1}^3 q_i(\xi)   }
\end{align}
and $\alpha_i = q_i(\bar{\lambda})$.
Indeed,
\begin{align*}
    \lambda & \asymp 
    \beta_T \beta_M \alpha_1 \alpha_2 \alpha_3 + \frac{ \beta_T \alpha_3 }{ \sqrt{n_M} } \sum_{ij} \mathbb{E}[ \evu_i \evv_j \ermZ_{ij} ] + \frac{1}{\sqrt{n_T}} \sum_{ijk} \mathbb{E}[ \evu_i \evv_j \evw_k \etW_{ijk} ] \\
    & = \beta_T \beta_M \alpha_1 \alpha_2 \alpha_3 + \frac{ \beta_T \alpha_3 }{ \sqrt{n_M} } \sigma_M^2 \sum_{ij} \mathbb{E}[ \frac{\partial \evu_i}{\partial \ermZ_{ij}} \evv_j ] + \frac{ \beta_T \alpha_3 }{ \sqrt{n_M} } \sigma_M^2 \sum_{ij} \mathbb{E}[ \evu_i \frac{\partial \evv_j}{\partial \ermZ_{ij}} ] \\
    & \quad \quad + \frac{\sigma_T^2}{\sqrt{n_T}} \sum_{ijk} \mathbb{E}[ \frac{\partial \evu_i}{\partial \etW_{ijk}} \evv_j \evw_k ] + \frac{\sigma_T^2}{\sqrt{n_T}} \sum_{ijk} \mathbb{E}[ \evu_i \frac{\partial \evv_j}{\partial \etW_{ijk}} \evw_k ] + \frac{\sigma_T^2}{\sqrt{n_T}} \sum_{ijk} \mathbb{E}[ \evu_i \evv_j \frac{\partial \evw_k}{\partial \etW_{ijk}} ] \\
    & \asymp \beta_T \beta_M \alpha_1 \alpha_2 \alpha_3 - \frac{\beta_T^2 \alpha_3^2 }{ n_M} \sigma_M^2 \sum_{ij} \mathbb{E}[ \ermG^{11}_{ii} \evv_j^2 ]
    - \frac{\beta_T^2 \alpha_3^2 }{ n_M} \sigma_M^2 \sum_{ij} \mathbb{E}[ \ermG^{22}_{jj} \evu_i^2 ] \\
    & \quad \quad - \frac{\sigma_T^2 }{ n_T} \sum_{ijk} \mathbb{E}[ \ermG^{11}_{ii} \evv_j^2 \evw_k^2 ] - \frac{\sigma_T^2 }{ n_T} \sum_{ijk} \mathbb{E}[ \ermG^{22}_{jj} \evu_i^2 \evw_k^2 ] - \frac{\sigma_T^2 }{ n_T} \sum_{ijk} \mathbb{E}[ \ermG^{33}_{kk} \evu_i^2 \evv_j^2 ] \ ,
\end{align*}
which provides:
\begin{equation}
    \label{eq:lambda}
    \boxed{
    \bar{\lambda} = \beta_T \beta_M \alpha_1 \alpha_2 \alpha_3 - \bar{\gamma} \sigma_M^2 (g_1(\bar{\lambda}) + g_2(\bar{\lambda})) - \sigma_T^2 g(\bar{\lambda})
    } \ .
\end{equation}

\subsection{Proof of Proposition \ref{prop:performance_clustering}}
The proof is a straightforward application of Theorem \ref{thm:stat_unit} to the multi-view model in Eq. (\ref{eq:multiview}). In fact, given $\hat \vy$ the 2-mode singular vector of $\tX$ corresponding to its largest singular value, we then have by Theorem \ref{thm:stat_unit}:
\begin{align}
    \vert \langle \hat \vy, \bar \vy \rangle \vert \asto \alpha = q_2(\bar \lambda),
\end{align}
where $\bar \lambda$ and $q_2(\cdot)$ are defined in Proposition \ref{prop:performance_clustering}. Moreover, the vector $\hat \vy$ decomposes as:
\begin{align}
    \hat\vy = \alpha \bar \vy + \sigma \bar \vy^{\perp},
\end{align}
where $\bar \vy^{\perp} \in \sR^n$ is a random vector, orthogonal to $\bar \vy$ and of unit norm. Since $\bar \vy$ is also of unit norm, $\sigma$ satisfies $1 = \alpha^2 + \sigma^2$, therefore $\sigma = \sqrt{1 - \alpha^2}$. Finally, the Gaussianity of the entries of $\hat \vy$ can be obtained thanks to similar arguments as in \citep{couillet2016kernel}.
\section{Additional Simulations}\label{appendix:simulations}
\begin{figure}[h!]
    \centering
    \input{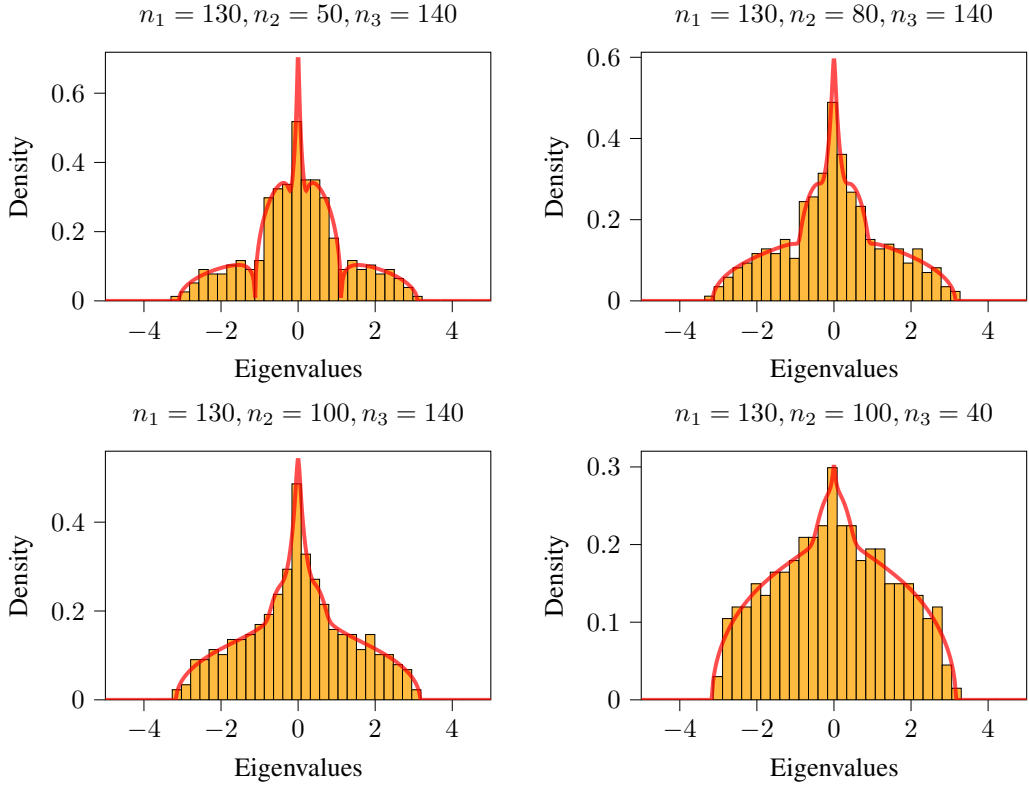}
    \caption{Empirical versus limiting spectrum of the random matrix $\mH$ varying the tensor dimensions $n_i$ and $\beta_T = 2, \beta_M=3$. The limiting measure is computed using Algorithm \ref{alg:stieltjes_transform} which implements the result of Theorem \ref{thm:stat_unit}.}
    \label{fig:specturm}
\end{figure}

In this section, we provide further simulations to support our findings. Precisely, we depict in Figure \ref{fig:specturm} histograms representing the empirical spectral measure of the random matrix $\rmH$ along with the limiting spectral measure obtained in Theorem \ref{thm:spectrum_unit}, varying the dimensions of the tensor $n_i$'s. We can notice that the empirical histograms are accurately captured by the limiting measure $\nu$. Moreover, Figure \ref{fig:alignments_appendix} depicts the asymptotic summary statistics as per Theorem \ref{thm:stat_unit} with the corresponding simulated ones. Again, our results seem to capture well the empirical behavior. Importantly, as we discussed in Section \ref{sec:summary_stats}, in the setting of Figure \ref{fig:alignments_appendix} ($\beta_T = 1$), our results predict a discontinuous behavior for the alignments $\alpha_1, \alpha_2$ which is typical for spiked random tensors \citep{jagannath2020statistical}. 

\begin{figure}[h!]
    \centering
    \input{figs/alignments_appendix}
    \caption{Empirical versus asymptotic summary statistics for $n_1 = 40, n_2 = 110, n_3 = 90, \beta_T = 1$ and varying $\beta_M$. Simulations are obtained by averaging over $10$ independent realizations of the random matrix $\rmZ$ and tensor $\tW$. Our results exhibit a phase transition when varying $\beta_M$ above which the matrix components $(\vx, \vy)$ become estimable.}
    \label{fig:alignments_appendix}
\end{figure}


\end{document}